\documentclass[10pt,twocolumn,letterpaper]{article}

\PassOptionsToPackage{numbers, compress}{natbib}

\usepackage[pagenumbers]{cvpr}

\usepackage[utf8]{inputenc} 
\usepackage[T1]{fontenc}    
\usepackage[pagebackref,breaklinks,colorlinks,allcolors=cvprblue]{hyperref}
\usepackage{url}            
\usepackage{booktabs}       
\usepackage{amsfonts}       
\usepackage{nicefrac}       
\usepackage{microtype}      
\usepackage{xcolor}         
\usepackage[accsupp]{axessibility}

\newcommand{\parsection}[1]{\vspace{2pt}\noindent\textbf{#1.}}

\usepackage{multirow}
\usepackage{textcomp}
\usepackage{tabulary}
\usepackage{colortbl}
\usepackage{siunitx}
\usepackage{subcaption}

\newcommand{\sbf}[1]{\bfseries #1}

\usepackage{graphicx}
\usepackage{nicematrix}
\definecolor{tabfirst}{rgb}{1, 0.7, 0.7} 
\definecolor{tabsecond}{rgb}{1, 0.85, 0.7} 
\definecolor{tabthird}{rgb}{1, 1, 0.7} 

\usepackage{gensymb}
\usepackage{mathtools}

\def\modelname{{R3D2}}
\def\datasetname{{R3D3}}

\usepackage[symbol]{footmisc}

\definecolor{CornflowerBlue}{RGB}{100, 149, 237}
\definecolor{YellowGreen}{RGB}{154, 205, 50}
\definecolor{Apricot}{RGB}{251, 206, 177}

\usepackage[capitalize]{cleveref}
\crefname{section}{Sec.}{Secs.}
\Crefname{section}{Section}{Sections}
\Crefname{table}{Table}{Tables}
\crefname{table}{Tab.}{Tabs.}

\makeatletter
\DeclareRobustCommand\onedot{\futurelet\@let@token\@onedot}
\def\@onedot{\ifx\@let@token.\else.\null\fi\xspace}

\makeatother

\definecolor{cvprblue}{rgb}{0.21,0.49,0.74}

\def\confName{CVPR}
\def\confYear{2026}

\title{{\modelname}: \underline{R}ealistic \underline{3D} Asset Insertion via \underline{D}iffusion for Autonomous Driving Simulation}

\author{%
  William Ljungbergh$^{*,1,2,\dagger}$
  \and
  Bernardo Taveira$^{*,1, 3}$
  \and
  Wenzhao Zheng$^{4}$
  \and
  Adam Tonderski$^{1}$
  \and
  Chensheng Peng$^{4}$
  \and
  Fredrik Kahl$^{3}$
  \and
  Christoffer Petersson$^{1, 3}$
  \and
  Michael Felsberg$^{2}$
  \and
  Kurt Keutzer$^{4}$
  \and
  Masayoshi Tomizuka$^{4}$ 
  \and
  Wei Zhan$^{4}$
  \and
  $^1$Zenseact \quad $^2$Linköping University \quad $^3$Chalmers University \quad $^4$UC Berkeley
}
  
\begin{document}

\twocolumn[{
\maketitle
\vspace{-10mm}
\begin{center}
    \captionsetup{type=figure}
    \includegraphics[width=1.0\textwidth,trim={0cm 0cm 0cm 0cm},clip]{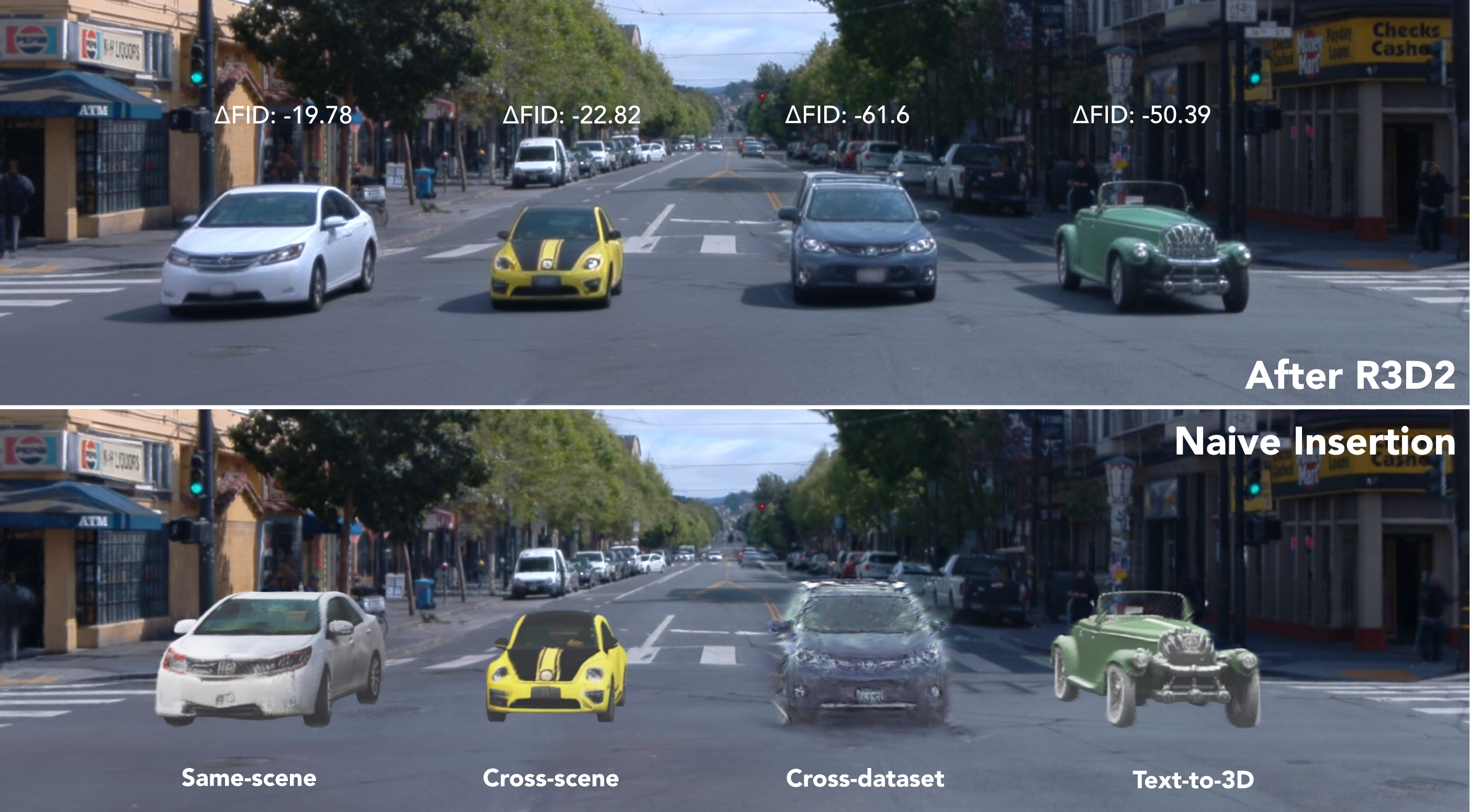}
    \captionof{figure}{{\modelname} enables realistic insertion of 3D assets from different origins into existing scene reconstructions. From left to right, assets are sourced from: the same scene, a different scene, a different dataset, and a text-to-3D generative model.}
    \label{fig:front-fig}
\end{center}
}]

\begingroup
\renewcommand{\thefootnote}{\fnsymbol{footnote}}
\footnotetext[1]{Equal contribution. $^{\dagger}$Work done while at UC Berkeley.}
\endgroup

\begin{abstract}
Validating autonomous driving (AD) systems requires diverse and safety-critical testing, making photorealistic virtual environments essential. Traditional simulation platforms, while controllable, are resource-intensive to scale and often suffer from a domain gap with real-world data. In contrast, neural reconstruction methods like 3D Gaussian Splatting (3DGS) offer a scalable solution for creating photorealistic digital twins of real-world driving scenes. However, they struggle with dynamic object manipulation and reusability as their per-scene optimization-based methodology tends to result in incomplete object models with integrated illumination effects. This paper introduces {\modelname}, a lightweight, one-step diffusion model designed to overcome these limitations and enable realistic insertion of complete 3D assets into existing scenes by generating plausible rendering effects—such as shadows and consistent lighting—in real time. This is achieved by training {\modelname} on a novel dataset: 3DGS object assets are generated from in-the-wild AD data using an image-conditioned 3D generative model, and then synthetically placed into neural rendering-based virtual environments, allowing {\modelname} to learn realistic integration. Quantitative and qualitative evaluations demonstrate that {\modelname} significantly enhances the realism of inserted assets, enabling use-cases like text-to-3D asset insertion and cross-scene/dataset object transfer, allowing for true scalability in AD validation. To promote further research in scalable and realistic AD simulation, we release our code, see {\color{teal}\href{https://research.zenseact.com/publications/R3D2/}{project page}}. 
\end{abstract}

\section{Introduction}
Ensuring safety remains one of the most important---if not the most important---aspects of autonomous driving (AD) development.
To achieve this, autonomous vehicles (AVs) must be validated across a wide range of scenarios, covering environmental variations as well as diverse road-user appearances and behaviors.
However, validating AVs at this scale in the real world is costly, time-consuming, and potentially dangerous in the case of safety-critical testing.
Therefore, virtual environments have emerged as a crucial tool for testing and validating AVs in a safe, controlled, and scalable manner.

Although traditional simulators (e.g., CARLA~\cite{dosovitskiy2017carla}) offer high controllability, the creation of environments and assets for these platforms is resource-intensive and often results in a domain gap with real-world AV data.
In remedy, neural rendering has emerged as a scalable alternative for creating highly realistic and controllable digital clones of real world driving data.
Recent advances in Neural Radiance Fields (NeRFs)~\cite{mildenhall2021nerf} and 3D Gaussian Splatting (3DGS)~\cite{kerbl20233d} have demonstrated the ability to efficiently reconstruct driving scenes with minimal sim-to-real gap~\cite{lindstrom2024nerfs}.
Such reconstructions can be used to create a large amount of safety-critical scenarios~\cite{ljungbergh2025neuroncap} to validate a full-stack AD system.

Despite achieving impressive reconstruction quality, current neural rendering techniques like 3DGS still rely on per-scene optimization, which limits their ability to generate truly diverse and novel scenarios.
For instance, road users are typically observed from limited viewpoints in driving data, leading to incomplete or partially reconstructed 3D models.
Consequently, if these actors are moved or viewed from novel perspectives not present in the original recordings, significant visual artifacts can emerge.
Furthermore, environmental effects such as illumination and shadows are often integrated into these partial reconstructions, restricting the reuse of assets in different scenes or under varying lighting conditions.
Addressing these challenges could further enhance the scalability and versatility of neural rendering techniques in generating truly novel scenarios.

Replacing foreground objects with high-quality 3D assets offers a promising path toward overcoming the aforementioned limitations of per-scene neural reconstructions. 
Yet, without scene-specific illumination and shadow effects, these assets often fail to integrate realistically into the environments they are placed in.
Physically-based rendering (PBR) with ray tracing provides a principled solution to this challenge, but requires detailed knowledge of material properties (e.g., albedo, roughness, specularity) and scene lighting (e.g., source intensity, color, and direction)—information that is typically unavailable in neural reconstruction pipelines. 
Although there are works that seek to recover such parameters \cite{wang2022neural, liang2024dipir}, the problem remains fundamentally ill-posed and error-prone given the sparse, incomplete observations common in driving data. 

We explore an alternative approach---leveraging large generative models (LGMs) as priors to bypass the need for accurate scene-level modeling and physical attribute estimation while still enabling realistic insertion of assets into virtual environments. LGMs, when trained on internet-scale real world data, have proven effective in various applications including image in-painting~\cite{rombach2022high}, text-to-3D generation~\cite{xiang2024structured}, and in improving the fidelity of neural rendering-based reconstruction methods~\cite{Liu20243dgsenhancer}. However, they struggle at our particular task, either not maintaining overall scene consistency or failing to generate realistic rendering effects. To overcome these limitations, we propose {\modelname}, a lightweight, one-step diffusion model trained to enable realistic insertion of foreign 3D assets into existing scene reconstructions by generating plausible rendering effects like shadows and lighting.

To train {\modelname}, we construct a novel dataset of 3DGS object assets derived from in-the-wild AD data using an image-conditioned large generative 3D model~\cite{xiang2024structured}. 
While these assets closely match the shape and appearance of its corresponding real-world objects, they initially lack realistic integration when placed into new scenes.
We then leverage state-of-the-art AD scene reconstruction methods~\cite{hess2024splatad} to capture the original environment, replacing existing objects with our generated 3DGS assets.
This process yields training pairs: scenes with realistically shaped and placed assets that still exhibit a sim-to-real gap in rendering.
{\modelname} is then trained to bridge this gap by learning the distribution of rendering effects from the original scenes, effectively realizing photorealistic inserted 3DGS assets.
Beyond quantitative evaluations, we demonstrate {\modelname}'s utility through highly useful applications, including text-to-3D asset insertion and the flexible transfer of objects between different scenes and even datasets. Our contributions can be summarized as follows:
\begin{itemize}
    \item Propose a lightweight diffusion model, {\modelname}, capable of generating rendering effects in real time, drastically improving realism of inserted assets.
    \item Demonstrate powerful asset manipulation capabilities while also generalizing to novel applications, including text-to-3D generation and seamless cross-dataset asset insertion.
    \item Construct the novel dataset used to train R3D2, comprising of image pairs where the original actors are replaced with the aforementioned assets.

\end{itemize}

\section{Related Work}
\label{sec:related_work}
\parsection{Simulation for autonomous driving}
Simulation is indispensable for the development and testing of autonomous driving systems.
Conventional methods~\cite{dosovitskiy2017carla, shah2018airsim} achieve high visual fidelity via Physically-Based Rendering (PBR) and ray tracing but rely on manually designed 3D assets and environments.
Neural Radiance Fields (NeRFs)~\cite{mildenhall2021nerf} and its adaptations to autonomous driving~\cite{turki2023suds,unisim,tonderski2024neurad,yang2024emernerf} bypass the need for manual modeling by enabling the generation of photorealistic 3D representations from real-world data.
Compared to NeRF's ray-tracing formulation, 3D Gaussian Splatting (3DGS)~\cite{kerbl20233d} uses rasterization to increase the rendering speed by orders of magnitude, allowing real-time simulation of full AD sensor suites~\cite{yan2024street, zhou2024drivinggaussian, hess2024splatad}.
While offering a scalable solution for generating realistic virtual environments, neural-rendering-based reconstructions often result in partial or incomplete reconstructions of the dynamic objects due to the sparse nature of AD data. 
This incomplete modeling impedes object manipulation, a crucial capability of a simulator, and alludes to the need for better object reconstruction and/or the ability to insert novel and complete objects into the scene, the latter of which is the focus of this work.

\parsection{Diffusion models and 3D asset generation}
Diffusion models~\citep{ho2020denoising,rombach2022high} have emerged as a cornerstone for modern generative modeling due to their ability to produce high-quality samples.
Following the introduction of latent diffusion models~\cite{rombach2022high}, 2D image generation methods have largely focused on two fronts: scaling models to internet-scale datasets to achieve unprecedented photorealism~\cite{ramesh2022hierarchicaltextimagegeneration, imagen, esser2024sd3} and enhancing sampling efficiency and speed~\citep{sauer2024adversarial, chen2025sana-sprint}.
Building on the success in 2D, the field has progressively extended these powerful generative capabilities to 3D asset creation.
TRELLIS3D~\cite{xiang2024structured} employs a structured latent representation to generate detailed 3D objects with intricate geometry and textures, supporting multiple output formats including radiance fields, 3D Gaussians, and meshes.
Concurrently, Amodal3R~\cite{wu2025amodal3r} builds upon similar structured representations to tackle amodal 3D object reconstruction from a single image. This approach aims to accurately infer the complete 3D shape, rotation, and translation of objects, even when partially occluded, demonstrating progress towards understanding 3D scenes from limited visual input.

\parsection{Asset insertion and relighting}
Seamless object insertion into foreign scenes, particularly achieving consistent relighting, remains a prominent challenge in computer vision and graphics. Physically-Based Rendering (PBR) techniques \cite{pharr2023physically} offer realistic relighting but demand detailed 3D assets and material properties, constraining practical applications.
Other recent hybrid diffusion-PBR methods like DiPIR~\cite{liang2024dipir} remain computationally demanding in inference and necessitate asset material knowledge.
To circumvent these requirements, generative relighting methods such as IC-Light~\cite{zhang2025scaling} adapt 2D object appearance to new scenes but often neglect reciprocal scene modifications like shadow casting.
Conversely, other methods~\cite{winter2024objectdrop, dhiman2025reflecting} focus on scene adaptation for object integration (e.g., plausible shadow generation) but do not alter the inserted object's appearance to match the new environment. Consequently, simultaneous bidirectional object-scene adaptation is non-trivial.

While general-purpose image editing has advanced with powerful generative models \cite{meng2022sdedit, Brooks_2023_CVPR, huang2024smartedit}, these often exhibit limitations, such as difficulties with high-resolution synthesis \cite{meng2022sdedit} or maintaining consistency with the source image \cite{meng2022sdedit, Brooks_2023_CVPR}, as shown by \cite{parmar2024onestepimagetranslationtextetoimag}.
Pix2Pix-Turbo \cite{parmar2024onestepimagetranslationtextetoimag} demonstrated enhanced fidelity preservation, yet its framework is typically tailored to specific, predefined image-to-image translation tasks. Difix3D+ \cite{wu2025difix3d} takes this architecture and shows the potential of it when trained to fix gaussian artifacts. Our work builds upon the foundational principles of Pix2Pix-Turbo, adapting its architecture for the nuanced task of seamless object integration. The primary goal is to achieve holistic visual consistency by concurrently modifying both the object's appearance to match scene illumination and context, and the surrounding scene (e.g., rendering shadows) to realistically incorporate the new object, ensuring mutual coherence.

\section{Method}
In this section we present {\modelname}, a lightweight diffusion model trained to realistically integrate 3D assets into existing scene reconstructions without the need for explicit material attributes, surface normals or environment maps.
To achieve this, we first construct a specialized training dataset, named {\datasetname}, derived from in-the-wild autonomous driving data.
This dataset comprise image pairs that can be used to train {\modelname} to learn complex rendering effects like shadows and lighting adjustments.
We outline our dataset generation process in \cref{sec:dataset_construction}, and then describe the architecture of {\modelname} and its training process in \cref{sec:diffusion_model}.

\begin{figure}[!t]
    \centering
    \includegraphics[trim={0 0 0 0},clip,width=\linewidth]{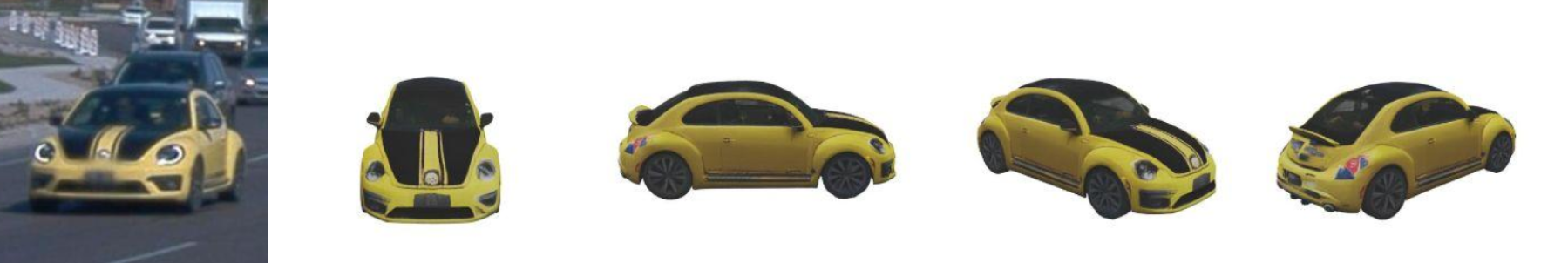}
    \caption{Complete 3D asset generated from an in-the-wild object observation using Amodal3R~\cite{wu2025amodal3r}.}
    \label{fig:asset-generation}
\end{figure}

\begin{figure}[!t]
    \centering
    \includegraphics[width=\linewidth]{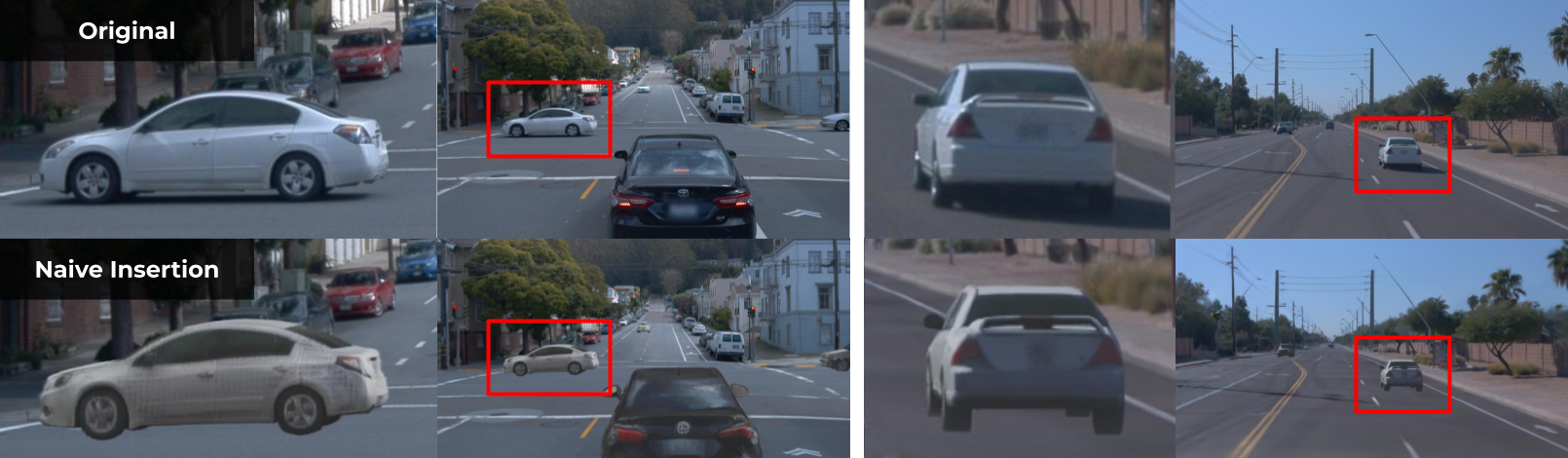}
    \caption{Two image pairs from {\datasetname}. The top row shows the original image $I_\text{target}$ and the bottom row shows the input image $I_\text{input}$. Inserted actors have been enlarged for visualization purposes.}
    \label{fig:training_pair_example}
\end{figure}

\begin{figure*}[!t]
    \centering
    \includegraphics[width=\linewidth]{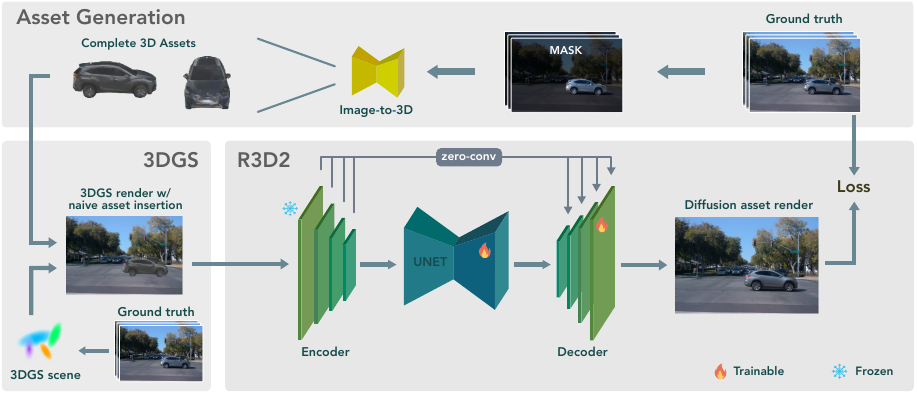}
\caption{Overview of the {\modelname} training pipeline. Amodal3R~\cite{wu2025amodal3r} generates assets from the original data, which are then inserted into 3DGS-reconstructed environments~\cite{hess2024splatad} in place of the corresponding real objects. Rendering the modified scene alongside the original yields training pairs, enabling {\modelname} to learn to add realistic shadows and lighting effects.}
    \label{fig:method_diagram}
\end{figure*}

\subsection{{\datasetname}: Dataset for training {\modelname}}
\label{sec:dataset_construction}

Training {\modelname} requires image pairs depicting the same scene: one with a synthetically inserted 3DGS asset that lacks realistic environmental integration ($I_\text{input}$), and a corresponding photorealistic target image derived from the original scene ($I_\text{target}$).
We construct this dataset using the Waymo Open Dataset (WOD)~\cite{waymo} and its object centric extension (WOD-OA)~\cite{shen2023gina3d}.
WOD provides extensive driving sequences with raw sensor data, while WOD-OA offers instance-aligned image crops and segmentation masks for objects within these scenes.
The construction of {\datasetname} involves three key stages: asset generation, scene reconstruction with asset re-insertion, and data curation.

\begin{figure*}[t]
     \footnotesize
     \centering
     \includegraphics[width=1.0\textwidth]{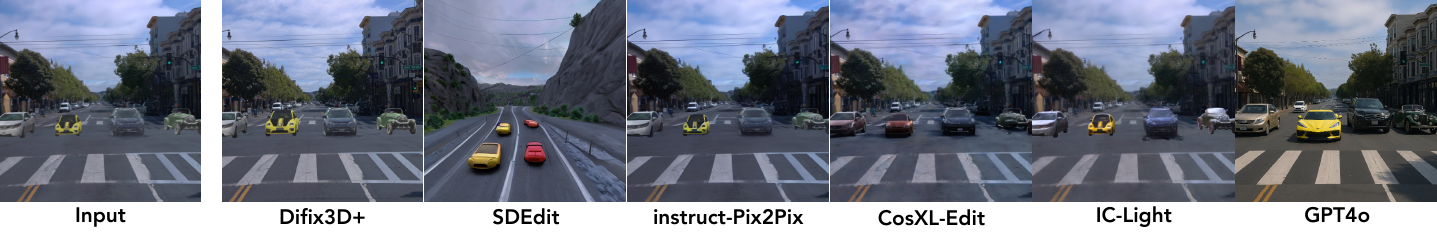} 
     \caption{Image editing results when applying several SotA image-editing tools to our setting. {\modelname} performance on this same image is shows in \cref{fig:front-fig}.} 
     \label{fig:baselines_failure_compact}
\end{figure*}

\parsection{High-Fidelity 3D Asset Generation}
To create complete and reusable 3D assets, we employ Amodal3R~\cite{wu2025amodal3r}, an image-conditioned 3D generation model.
As illustrated in \cref{fig:asset-generation}, Amodal3R takes an object-centric image crop, its corresponding 2D segmentation mask, and a foreground occlusion mask as input to produce a textured 3D asset.
While WOD-OA directly provides the image crops and initial segmentation masks, the foreground occlusion masks (indicating parts of the object occluded by other foreground elements) are not available.
We generate these by estimating a complete amodal segmentation mask for the target object using pix2gestalt~\cite{ozguroglu2024pix2gestalt}.
The difference between this amodal mask and the original segmentation mask from WOD-OA yields the required foreground occlusion mask.
Further details on the foreground occlusion mask generation process are provided in the supplementary material. Each generated asset is represented as a set of 3D Gaussian primitives.

\parsection{Scene reconstruction and asset re-insertion}
With a collection of 3DGS assets, we proceed to generate the training image pairs. We select approximately 300 sequences from WOD deemed suitable for high-quality neural reconstruction, see full details in supplementary material.
For each sequence, we reconstruct a virtual environment using a state-of-the-art neural reconstruction method, SplatAD~\cite{hess2024splatad}, with minor modifications detailed in the supplementary material.
SplatAD decomposes the scene into a static background and dynamic actors, each represented by a set of Gaussian primitives, allowing us to easily replace dynamic actors with our assets.

To create a training pair, we first identify dynamic actors in the original scene and remove their corresponding Gaussian primitives from the scene reconstruction. Notably, for moving objects this also removes the corresponding shadows.
In their place, we insert their corresponding generated 3DGS assets aligning it to the pose (position and orientation) of the original, removed actors.
Rendering this modified scene (static background + new 3DGS assets) yields the input image $I_\text{input}$.
This image contains the geometrically correct assets but often exhibits a visual mismatch partly due the lack of scene-specific shadow casting, as exemplified in \cref{fig:training_pair_example}.
We can then use the original image to form a single training sample $(I_\text{input}, I_\text{target})$ for {\modelname}.

\parsection{Data curation}
While Amodal3R~\cite{wu2025amodal3r} achieves impressive generation quality, conditioning on only 2D image inputs occasionally lead to imperfect asset geometries or appearances.
To ensure the quality of {\datasetname}, we perform a two-stage filtering process for the generated assets.
First, we automatically filter out assets whose 3D shape deviates significantly from the original object dimensions defined by the annotated 3D bounding boxes from WOD~\cite{waymo}.
The remaining assets undergo manual inspection, and any that appear unrealistic, do not visually correspond to the original object, or exhibit significant rendering artifacts are removed, yielding a total of 5071 assets (4014 from training sequences, 1057 from validation sequences).
These curated assets and their corresponding scene re-insertions form the foundation for training {\modelname} to perform realistic rendering refinement.

\subsection{\modelname: Diffusion for realistic rendering of inserted actors}
\label{sec:diffusion_model}

We formulate the problem of realistic object insertion as an image editing task. As described in \cref{sec:dataset_construction}, the input image $I_\text{input}$ is a rendering of a scene with inserted 3DGS assets, and the target image $I_\text{target}$ is the original image with the original objects. {\modelname} is tasked with refining $I_\text{input}$ to match $I_\text{target}$ by adding realistic rendering effects like shadows and lighting adjustments.

To achieve this, we employ a one-step diffusion model, enabling efficient inference and thus wider practical applicability, and train it to directly transform the input image $I_\text{input}$ to the target image $I_\text{target}$. We build our method on top of SD-Turbo~\cite{sauer2024adversarial}, a distilled text-to-image model, but discard its text-conditioning and use it in an image-to-image fashion. We perform a single backward diffusion step directly from the degraded $I_\text{input}$, without applying additional noise~\cite{parmar2024onestepimagetranslationtextetoimag}. This proves to be effective as $I_\text{input}$, although degraded with unrealistic objects, still contains a lot of information about the scene in terms of illumination and overall geometry. Compared to other image-to-image methods which concatenate the conditioning image with the denoise process \cite{Brooks_2023_CVPR, Sheynin2023EmuEP, Saharia2022palette}, our approach better preserves details in $I_\text{input}$ and does not require extensive fine-tuning, as shown in \cite{parmar2024onestepimagetranslationtextetoimag}. The model architecture and training process are illustrated in \cref{fig:method_diagram}.

\parsection{Variational autoencoder}
The choice of using a distilled single-step diffusion process is key to real-time speed and has been shown to yield good performance \cite{sauer2024adversarial}. 
However, in this formulation, the autoencoder becomes a bottleneck for inference speed.
Therefore, we instead use a distilled version of the autoencoder~\cite{madebyollinTAESD}, and ablate this tradeoff in \cref{sec:runtime}.
In addition, we find that details of the input image are inevitably lost when encoding it to the latent space, and following \cite{parmar2024onestepimagetranslationtextetoimag, Liu20243dgsenhancer, wu2025difix3d} we address this by adding a skip connections from the encoder to the decoder.

\parsection{Fine-tuning strategy}
Training is performed similarly to pix2pix-Turbo~\cite{parmar2024onestepimagetranslationtextetoimag}, i.e., by consolidating the VAE Encoder, UNet, and VAE Decoder into a single end-to-end trainable block.
The single-step nature of the model enables this simplification of the training process without running into hardware limitations due to multiple passes through the UNet.
Given that the model is trained to refine the input image, opposed to generating a new image from pure noise, we adopt a lower initial conditional noise step as explored in~\cite{wu2025difix3d}.
We freeze the VAE encoder, fine-tune the UNet and VAE Decoder using LoRA~\cite{hu2022lowrank} for efficiency, and initialize the skip connections to zero.
The training is performed at a resolution of $1080\times1920$, significantly larger than the pretrained model ($512\times512$).


Training in an end-to-end fashion additionally facilitates the use of perceptual and style losses. Considering potential pixel-level misalignment between ground truth images and images with reinserted objects, we discard the pixel-aligned $L_2$ loss and instead use the perceptual LPIPS~\cite{zhang2018unreasonable} loss  as it is robust to small shifts and emphasizes higher-level structures. 
In addition, to encourage the model to learn fine details and produce sharper images, we employ a Gram matrix loss~\cite{reda2022film}. For more details on architecture and training procedure, we refer to the supplementary material.

\begin{table*}[t]
     \centering
     \caption{Results from insertion of same-scene assets on the {\datasetname} validation set. We report common pixel, perceptual, and feature-level similarity metrics for full images and actor crops. FPS is evaluated on an NVIDIA RTX5090. $^*$Original reconstruction metrics are shown for reference.}
     \label{tab:image_quality}
     \resizebox{\textwidth}{!}{%
          \begin{tabular}{l|
                    S[table-format=2.2, round-mode=places, round-precision=2]
                    S[table-format=1.3, round-mode=places, round-precision=3]
                    S[table-format=1.3, round-mode=places, round-precision=3]
                    S 
                    S[table-format=1.3, round-mode=places, round-precision=3]
                    S 
                    S[table-format=1.3, round-mode=places, round-precision=3]
                    | 
                    S 
                    S[table-format=1.3, round-mode=places, round-precision=3]
                    S 
                    S[table-format=1.3, round-mode=places, round-precision=3]
               }
               \toprule

                                    & \multicolumn{7}{c|}{Full resolution} & \multicolumn{4}{c}{Actor-centric crops}                                                                                                                                                                                                                                           \\
               Method               & \text{FPS$\uparrow$}                 & \text{CLIP IS} $\uparrow$               & \text{DINO IS} $\uparrow$ & \text{FID} $\downarrow$ & \text{LPIPS} $\downarrow$ & \text{PSNR} $\uparrow$ & \text{SSIM}$\uparrow$ & \text{FID} $\downarrow$ & \text{LPIPS} $\downarrow$ & \text{PSNR}$\uparrow$ & \text{SSIM} $\uparrow$ \\ \midrule
               Orig. Recon.$^*$     & \text{-}                             & 0.97752                                 & 0.9772                    & 9.169                   & 0.177                     & 28.94                  & 0.896                 & 8.97                    & 0.13                      & 27.90                 & 0.88                   \\\midrule
               Naive insertion      & \text{-}                             & 0.9556                                  & 0.9685                    & 22.523                  & 0.2108                    & 24.4808                & \sbf{0.866}           & 30.0166                 & 0.29397                   & 19.383                & 0.6852                 \\
               $+$ {\modelname}     & \sbf{13.359}                         & \sbf{0.9809}                            & \sbf{0.9872}              & 10.8510                 & 0.1419                    & 25.294                 & 0.8620                & 10.574                  & 0.22029                   & 20.49682              & 0.7071                 \\
               $+$ {\modelname}-BIG & 4.067                                & 0.9790                                  & \sbf{0.9868}                    & \sbf{10.6916913}        & \sbf{0.13826}             & \sbf{25.556045532}     & 0.8600486             & \sbf{10.2445}           & \sbf{0.21546}             & \sbf{20.83576}        & \sbf{0.71571}          \\\bottomrule
          \end{tabular}
     }
     \vspace{-2mm}
\end{table*}

\section{Experiments}
\label{sec:experiments}
In this section, we conduct experiments to validate the effectiveness of using {\modelname} to achieve realistic insertion of 3DGS assets into reconstructed driving scenes.
We first validate our system by re-inserting actors at their original locations, enabling direct comparison with ground truth, and computing traditional image quality metrics (\cref{sec:realistic_insertion_gt}).
Next, we show the benefits of having access to complete 3D assets, allowing for more flexible asset manipulation while maintaining scene realism (\cref{sec:controllability}).
Then, we show how R3D2 enables the insertion of completely \textit{foreign} assets, sourced from other scenes, datasets, or even generated directly from text prompts (\cref{sec:foreign_object_insertion}).
Finally, we ablate our design choice of employing a smaller, distilled autoencoder ({\modelname}) versus the original, larger version ({\modelname}-BIG), and analyze the resulting trade-offs in runtime and image quality (see \cref{sec:runtime}).

\parsection{Metrics}
Where ground truth is available, we report standard image quality metrics, namely Peak Signal-to-Noise Ratio (PSNR), Structural Similarity Index (SSIM)~\cite{wang2004image}, Learned Perceptual Image Patch Similarity (LPIPS)~\cite{zhang2018unreasonable} as well as CLIP- and DINO-similarity scores~\cite{ruiz2023dreambooth}.
To evaluate realism without ground truth, we utilize Fréchet Inception Distance (FID)~\cite{heusel2017gans} computed on both full-resolution images (FID) and actor-centric crops (FID-A).

\parsection{Baselines}
As alluded to in \cref{sec:related_work}, prior work enabling realistic integration of 3D assets into novel scenes, without access to explicit lighting or material information, is scarce.
However, general image editing methods based on image/text prompts could--in theory--address this task. 
We evaluate the most promising candidates qualitatively in \cref{fig:baselines_failure_compact}.
Despite multiple prompt variations, SDEdit~\cite{meng2022sdedit} struggles to retain the content of the input image. Instruction based methods like InstructPix2Pix~\cite{Brooks_2023_CVPR} and CosXL-edit~\cite{podell2023sdxl} maintain consistency, but fail to produce realistic shadows and illumination effects for the inserted assets.
IC-Light~\cite{zhang2025scaling}, a model developed for the related task of 2D object relighting is by design unable to alter the environment (e.g., adding shadows), but surprisingly also struggles to realistically re-light the inserted assets.
Difix3D+~\cite{wu2025difix3d}, despite similar in architecture and close application also falls short when applied to our domain. It fixes most gaussian artifacts, saturates and sharpens the image, but doesn't attempt to add any shadows or fix any missing light effects.
While the state-of-the-art closed-access model GPT4o~\cite{openai2025image} is capable of producing highly realistic results with excellent lighting, it fails to preserve the original scene geometry and content. Further examples are provided in the supplementary material.
Therefore, we compare {\modelname} to naively inserting the 3DGS assets into the scene without any further processing (Naive Insertion), and contextualize it with respect to the original reconstruction method (Orig. Recon.), since the aforementioned baselines are not sufficiently capable to serve as meaningful points of comparison.

\subsection{Realistic re-insertion of same-scene assets}
\label{sec:realistic_insertion_gt}
\begin{figure*}[t]
     \footnotesize
     \centering
     \includegraphics[width=0.80\textwidth]{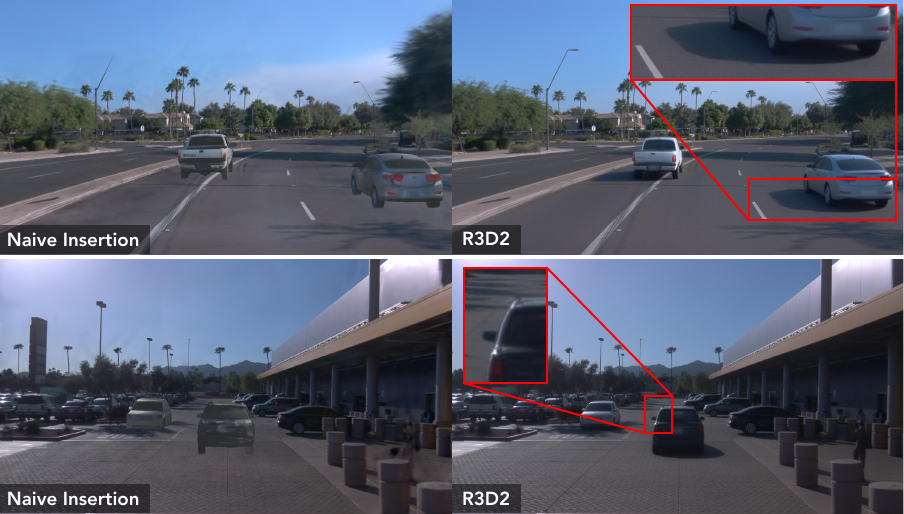} 
     \caption{Qualitative results from {\datasetname}. Realistically added effects from {\modelname} are highlighted in the images, such as the shadow on first example and the sun reflection on the second.}
     \label{fig:light_effects}
\end{figure*}

In \cref{tab:image_quality}, we evaluate the performance of {\modelname} on the task of inserting 3D assets into reconstructed driving scenes, computed both on full-resolution images and actor-centric crops provided by WOD-A~\cite{shen2023gina3d}.
We use the validation set of {\datasetname} (outlined in \cref{sec:dataset_construction}) and insert the asset generated from the same scene into the original reconstruction with the same poses as the original actors, allowing for direct comparison with the original image.
To contextualize the the results, we also show the metrics for the original reconstruction method (Orig. Recon.).
Note however, that the original reconstruction has been trained (per-scene overfitting) on these exact images which unsurprisingly leads to very high scores.
As shown, {\modelname} drastically reduce the gap in realism between the naive insertion and the original reconstruction.
This is further shown qualitatively in \cref{fig:light_effects}, where we show examples of effects generated by {\modelname}.
First, it is able to implicitly infer the global illumination source and realistically cast shadows caused by the inserted objects.
The second example shows how {\modelname} is able to generate realistic reflections from the sun off the reflective body of the car.

\subsection{Realistic same-scene asset manipulation}
\label{sec:controllability}

While neural rendering methods achieve impressive reconstruction results (see \cref{tab:image_quality}), they struggle with complete object reconstructions, causing rapid quality degradation even for minor rotations, as shown in \cref{tab:extrapolation}. Inserting complete 3D assets maintains stable quality across rotations, but suffers from all the aforementioned problems with realism. We compare both approaches by measuring FID and FID-A scores for rotations of 0$\degree$, 10$\degree$, 20$\degree$, and 180$\degree$ and show that {\modelname} dramatically increases quality in both cases, enabling flexible scenario editing either through original actors or, for more dramatic transformations, through completed assets.

\begin{table}[t]
     \centering
     \caption{Robustness to object manipulation. Measuring FID and FID-A scores when assets, original and inserted, are rotated by 0, 10, 20, and 180 degrees.}
     \label{tab:extrapolation}
     \resizebox{\linewidth}{!}{%
     \begin{tabular}{l|*{2}{S}|*{2}{S}|*{2}{S}|*{2}{S}}

          \toprule
                           & \multicolumn{2}{c|}{\text{Rotated 0\textdegree}} & \multicolumn{2}{c|}{\text{Rotated 10\textdegree}} & \multicolumn{2}{c|}{\text{Rotated 20\textdegree}} & \multicolumn{2}{c}{\text{Rotated 180\textdegree}}                                                                                                            \\
          Method           & \text{FID} $\downarrow$                          & \text{FID-A} $\downarrow$                         & \text{FID} $\downarrow$                           & \text{FID-A} $\downarrow$                         & \text{FID} $\downarrow$ & \text{FID-A} $\downarrow$ & \text{FID} $\downarrow$  & \text{FID-A}$\downarrow$ \\ \midrule
          Orig. Recon.     & 9.169                                            & 8.97                                              & 14.676                                            & 22.63                                             & 23.08                   & 44.556                    & 46.753                   & 83.1932                  \\
          $+$ {\modelname} & \sbf{6.314340114593506}                          & \sbf{5.4494781494140625}                          & \sbf{8.929129600524902}                           & \sbf{8.599021911621094}                           & \sbf{11.32935619354248} & \sbf{12.345247268676758}  & \sbf{17.797161102294922} & \sbf{24.26163673400879}        \\\midrule
          Naive insertion  & 22.523                                           & 30.016                                            & 23.9331                                           & 31.92                                             & 26.83                   & 36.282                    & 24.92346                 & 40.642                   \\
          $+$ {\modelname} & \sbf{10.851}                                     & \sbf{10.574}                                      & \sbf{12.204}                                      & \sbf{11.41}                                       & \sbf{14.7197}           & \sbf{13.983}              & \sbf{13.698}             & \sbf{14.3859}                  \\
          \bottomrule
     \end{tabular}%
     }
     \vspace{-2mm}
\end{table}

\begin{figure*}[t]
     \footnotesize
     \centering
     \includegraphics[width=1.0\textwidth]{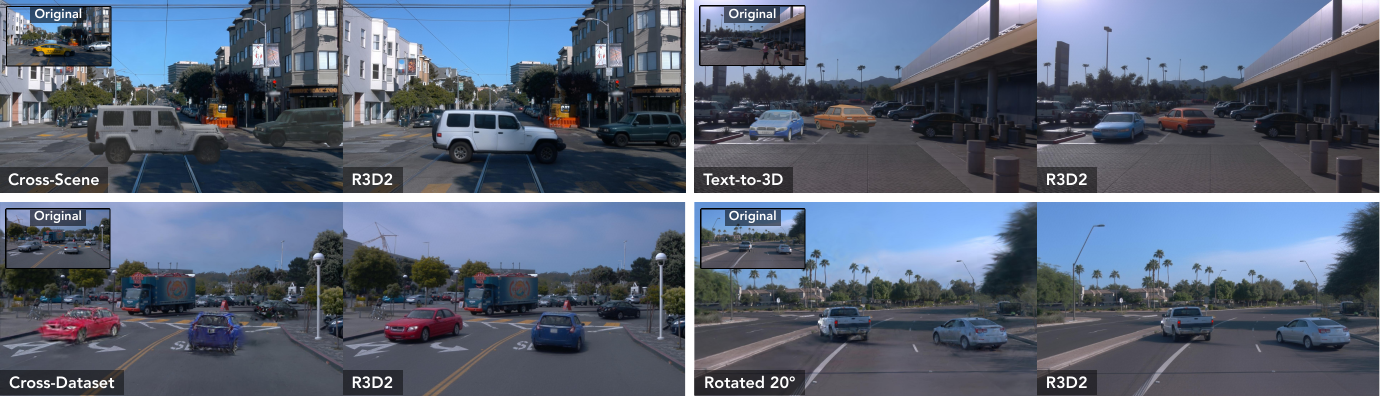}
     \caption{Qualitative examples of rendering effects generated with {\modelname}. We show results when rotating assets in the scene, as well as inserting assets from different sources, including other scenes in the same dataset, from other dataset, and text-to-3D generative models.}
     \label{fig:foreign_actors}
    \vspace{-2mm}
\end{figure*}

\subsection{Realistic insertion of foreign assets}
\label{sec:foreign_object_insertion}
The key advantage of our method is enabling the insertion of general assets into the virtual twins of real-world driving scenes.
To quantitatively study this capability we conduct a series of experiments where we insert \textit{foreign} assets (i.e., assets sourced from different scenes, different datasets, or generated from text prompts) into the driving scenes.
The goal of these experiments is to assess {\modelname}'s ability to generate realistic insertions of assets that are not native to the scene from which it was generated.
We conduct three experiments to evaluate the model's performance in this context: 1) cross-scene evaluation, where we replace the original actors with actors from other validation sequences; 2) cross-dataset evaluation, where we insert objects reconstructed from another AD dataset; and 3) text-to-3D evaluation, where we generate 3D assets from text prompts.

\begin{table}[t]
     \centering
     \caption{FID and FID-A scores demonstrating {\modelname}'s generalization to foreign object insertion: across scenes, across datasets (PandaSet~\cite{pandaset}), and with text-to-3D assets (TRELLIS~\cite{xiang2024structured}).}
     \label{tab:foreign_insertion}
     \resizebox{\linewidth}{!}{%
     \begin{tabular}{l|*{2}{S}|*{2}{S}|*{2}{S}}
          \toprule
                           & \multicolumn{2}{c|}{Cross-scene} & \multicolumn{2}{c|}{Cross-dataset} & \multicolumn{2}{c}{Text-to-3D}                                                                                   \\
          Method           & \text{FID} $\downarrow$           & \text{FID-A} $\downarrow$           & \text{FID} $\downarrow$        & \text{FID-A} $\downarrow$ & \text{FID} $\downarrow$ & \text{FID-A} $\downarrow$ \\ \midrule
          Naive insertion  & 26.078                            & 36.49                               & 46.420200347                   & 79.726402282              & 34.823                  & 66.23                     \\
          $+$ {\modelname} & \sbf{14.329}                      & \sbf{14.05}                         & \sbf{14.269}                   & \sbf{16.47999954223}      & \sbf{18.23}             & \sbf{15.853}                    \\
          \bottomrule
     \end{tabular}%
     }
\end{table}

\parsection{Cross-scene}
To evaluate the model's ability to generate realistic insertion of assets that are not native to the scene from which it was generated, we conduct a cross-scene evaluation.
More specifically, for a given validation sequence, we replace the all dynamic actors in the scene (similar to the dataset construction process) with actors from other validation sequences.
We match the actors to-be-replaced with actors from other sequences based on their size and randomly select a replacement actor that meets the size criteria.

\parsection{Cross-dataset}
To obtain a set of foreign assets, we reconstruct a small set of driving scenes from another dataset, namely PandaSet~\cite{pandaset}.
Specifically, we reconstruct 10 sequences commonly used for reconstruction~\cite{hess2024splatad} and extract the 3DGS representation of the all dynamic object present in the scenes.
As previously mentioned, the reconstruction is not always complete and we manually filter out object that are only partially reconstructed, see supplementary material for examples.
After filtering, we are left with 52 objects that are randomly inserted into the validation sequences of our dataset (similar to the cross-scene evaluation).

\parsection{Text-to-3D asset insertion}
To further demonstrate the versatility of our method, we show that our model can be used to insert objects generated from pure text into a driving scene.
This opens up a new avenue for realistically inserting assets that are rarely/never found in real-world recordings, further enabling the generation of diverse driving scenarios.
We randomly sample a set of prompts related to objects commonly found in driving scenarios (e.g., `a red SUV') and generate 3DGS representations of the prompts using TRELLIS~\cite{xiang2024structured}.
For full details of the prompts and the generated objects, we refer to the supplementary material.
In total, we generate $\sim400$ objects of various types and sizes and randomly replace original actors into the validation sequences. 

\parsection{Results}
For each of the three dataset perturbations, we compute FID and FID-A scores when using {\modelname}, compare it to naive insertion and show the results \cref{tab:foreign_insertion}.
As shown, {\modelname} is able to generate realistic insertions of foreign objects, with FID and FID-A scores that are similar to the same-scene insertion, shown in \cref{sec:realistic_insertion_gt}.
The results are further shown qualitatively in \cref{fig:foreign_actors}, where we replaced the original actors with assets from each of the three datasets.
Note that {\modelname} is not only able to cast shadows caused by the inserted object, but also to generate realistic shadows on the object itself (here caused by a shading tree in the top-left image).

\subsection{Real-time rendering}
\label{sec:runtime}

Our method is able to achieve realistic rendering effects without relying on computationally expensive PBR or ray tracing.
By employing a single-step diffusion process and a distilled autoencoder, our approach supports real-time inference—an important property for scaling simulation (see \cref{tab:image_quality}). Compared to {\modelname}-BIG, which retains the original autoencoder architecture, our method achieves a $3.3\times$ speedup in inference time, enabling real-time rendering of commonly used 10Hz cameras~\cite{alibeigi2023zenseact, caesar2020nuscenes, waymo}, with only a marginal reduction in perceptual and feature-level similarity metrics.

Performance benchmarks were conducted on consumer-grade hardware (NVIDIA RTX 5090) at a resolution of $1080\times1920$, without batching. Each benchmark was executed on a single GPU, using 1000 warmup steps followed by 1000 forward passes to compute the average inference time and standard deviation. We report the average inference time, as the standard deviation was negligible.
\section{Conclusion}
\label{sec:conclusion}
In this work, we presented {\modelname}, a lightweight and efficient diffusion-based model designed to enhance the realism of 3D asset insertion within existing 3D Gaussian Splatting (3DGS) representations of driving scenes.
By leveraging our novel dataset, {\datasetname}, constructed using 3DGS assets generated from in-the-wild AD data, {\modelname} learns to predict and apply plausible rendering effects, bridging the visual gap between inserted objects and the original scene. 
We demonstrated its efficacy through quantitative evaluations and qualitative results in highly useful applications, including text-to-3D asset insertion and flexible object transfer across scenes and datasets. {\modelname} offers a practical step towards creating more diverse, controllable, and high-fidelity simulation environments for the scalable and robust validation of autonomous driving 
systems.
Our work contributes to the development of safer autonomous vehicles, thereby offering positive societal impacts with limited potential for misuse.

\parsection{Limitations \& Future Work}
Our dataset creation process introduces alignment discrepancies between inserted assets and ground truth due to scale ambiguities and inaccuracies in the asset generation pipeline. Consequently, {\modelname} occasionally learns to replicate these errors, resulting in unintended modifications to geometry and placement in its outputs.
Additionally, the method does not address temporal consistency. When applied frame-by-frame to video sequences or across multi-camera rig setups, this can potentially lead to subtle flickering or inconsistencies over time or between different viewpoints. 
A natural solution would be building on video rather than image models, but current video models are too computationally intensive for real-time use. As more efficient architectures emerge, our dataset and training methodology should be directly applicable to achieve temporal and multi-view consistency.

\clearpage

\section*{Acknowledgements}
We thank Chenfeng Xu, Carl Lindström, and Georg Hess for valuable feedback.
This work was partially supported by Berkeley DeepDrive and the Wallenberg AI, Autonomous Systems and Software Program (WASP) funded by the Knut and Alice Wallenberg Foundation. Computational resources were provided by NAISS at \href{https://www.nsc.liu.se/}{NSC Berzelius}, and partially funded by the Swedish Research Council, grant agreement no. 2022-06725
{
    \small
    \bibliographystyle{ieeenat_fullname}
    \bibliography{main}
}

\clearpage
\appendix
\begin{center}
\Large
\textbf{Supplementary material}
\end{center}

\section{Implementation details}
\subsection{Model architecture}


As described in \cref{sec:diffusion_model}, {\modelname} builds upon the work of \cite{parmar2024onestepimagetranslationtextetoimag}, with key differences in the training procedure, application domain, and dataset. One of the most notable architectural changes to the base pretrained model lies in the autoencoder design. In particular, we modify the skip connections to capture the latent representations at three hierarchical levels of the encoder (or four in the case of {\modelname-BIG}), specifically just before each downsampling operation. These latent features are then processed using zero convolutions and subsequently added to the corresponding latent representations in the decoder after each upsampling stage. The zero convolutions are initialized as zeros during training.

\subsection{Training setup}

The pretrained model struggles with high resolution generation as it wasn't trained for it. When fine-tuned for our application at low resolutions the results are noticeably worse. Therefore, we train our model at a resolution of $1080 \times 1920$, the same resolution used for performing our inference tasks and benchmarks.

Given the high demands of training at high resolution we are limited to batch size of $1$ per GPU. In total $8\times$ A100 80GB GPUs are used for training in order to speed up development process. The training took about 10 hours, amounting to around 81 GPU hours.

We use the AdamW~\cite{loshchilov2017decoupled} optimizer with a learning rate of $lr=0.0001$ and employ a linear warm-up schedule for the first 500 steps. Our loss is a combination of the perceptual LPIPS loss~\cite{zhang2018unreasonable}, and the Gram-matrix loss~\cite{reda2022film}
\begin{equation*}
    \mathcal{L} = \lambda_{\text{LPIPS}}\mathcal{L}_{\text{LPIPS}} + \lambda_{\text{gram}} \mathcal{L}_{\text{gram}},
\end{equation*}
with $\lambda_{\text{LPIPS}} = 1.0$ and $\lambda_{\text{gram}} = 0.5$.
    
\subsection{Inference setup}

To evaluate the inference speed of the diffusion model with all our modifications we compile the models with the public available package \href{https://github.com/chengzeyi/stable-fast}{\texttt{stable-fast}}. Some of the optimizations performed by stable-fast are: fused CUDNN convolution kernels, fused GEMM kernels, fused multi-head attention and the use of CUDA graph format. All tests were performed at a resolution of $1080 \times 1920$ and at a batch size of one to make them relevant to real time data stream applications.

Tests were performed on the top consumer grade and enterprise grade GPUs as these should provide a more significant benchmark for years to come. As explained in \cref{sec:runtime}, $1000$ warm-up steps were performed followed by $1000$ forward passes through the model. This number was chosen by continuously running forwards passes and evaluating how both GPUs perform over time and how many steps they take to stabilize to a given speed. In our tests the NVIDIA H200 took the longest to stabilize and the value $1000$ was considered as a safe value to ensure proper warm-up for both. We performed the same test multiple times to ensure it resulted in negligible value changes. From the $1000$ forward passes we can compute the inference speed and its standard deviation. The results of the benchmark are shows in \cref{tab:full_speed_bench}.

\begin{table}[h]
     \centering
     \caption{Results for inference speed and deviation in frames per second (FPS) on both consumer hardware and enterprise server hardware.}
     \label{tab:full_speed_bench}
     \vspace{2mm}
     \begin{tabular}{l|cc}
          \toprule
                               & \multicolumn{2}{c}{Rendering speed (FPS)} \\
          Method               & \text{RTX 5090}            & \text{H200}      \\ \midrule
          Naive insertion      & \text{-}                   & \text{-}                             \\
          $+$ {\modelname}     & $13.359 \pm 0.024$         & $18.129 \pm 0.078$               \\
          $+$ {\modelname}-BIG & $4.067 \pm 0.005$          & $5.964 \pm 0.037$      \\
          \bottomrule
     \end{tabular}
    \vspace{-2mm}
\end{table}

\section{Shortcoming of current methods}

\subsection{Image-to-image style transfer baselines}
\begin{figure*}[ht]
    \centering
    \includegraphics[width=\linewidth]{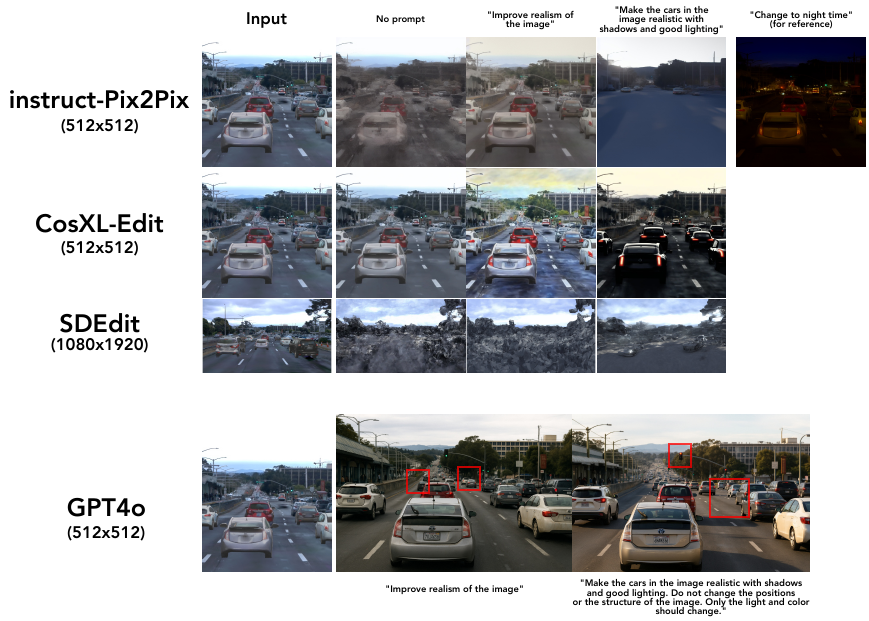}
\caption{Additional samples of off-the-shelf image editing models applied to our use case.}
    \label{fig:baselines_failure}
\end{figure*}

To further highlight the shortcomings of current image-to-image methods we provide extra samples with various prompts and different resolutions (\cref{fig:baselines_failure}). With GPT4o it is not possible to control the resolution of generation, but one can ask it to generate a square image instead which is the closes solution, however it drastically changes geometry and objects. SDEdit struggles to generate consistent images, particularly at higher resolution. Both instruct-Pix2Pix and CosXL-Edit fail to follow the directions and have a tendency to degrade the image further.

\begin{figure*}[t]
    \centering
    \vspace{6mm}
    \includegraphics[width=\linewidth]{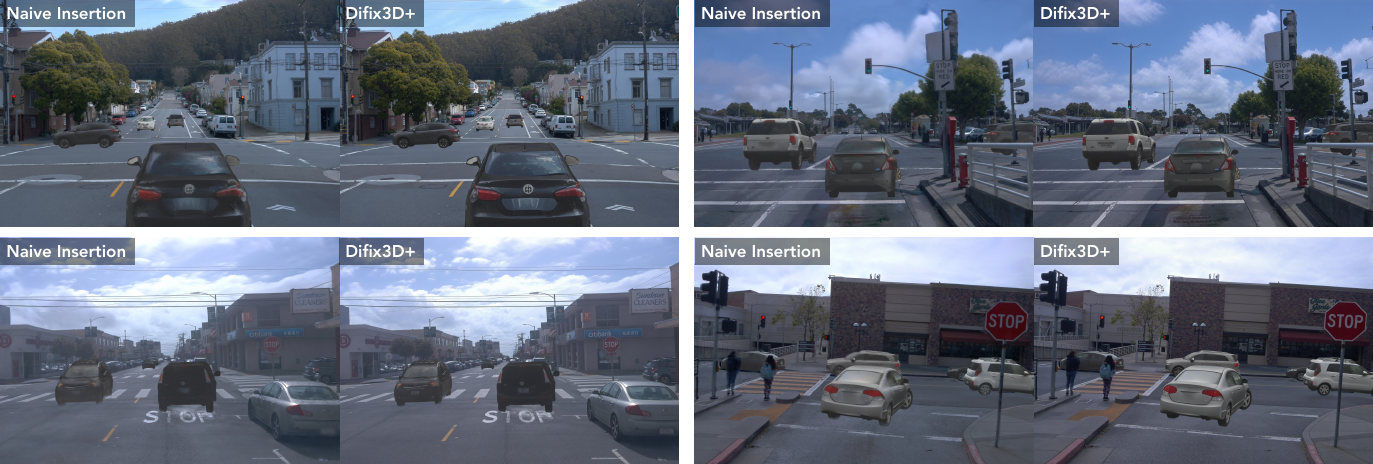}
\caption{Samples of Difix3D+ \cite{wu2025difix3d} applied on the R3D3 dataset. The model sharpens the image, saturates the colors, adds contrast and improves the look of the vehicle but it does not attempt to add any shadows or realistic light effects.}
    \label{fig:difix}
\end{figure*}

Given the similarities in methodology and application, we tested Difix3D+~\cite{wu2025difix3d} and showcase how foreign object insertion with realistic light effects cannot be generalized from training a model to fix gaussians (see \cref{tab:difix}). Some samples are provided in \cref{fig:difix}, demonstrating how this method is incapable of adding the shadows and light effects needed to blend the foreign asset to the environment.
\begin{table}[h]
    \centering
    \caption{Comparison between R3D2 and Difix3D+ on inserted assets}
    \label{tab:difix}
    \begin{tabular}{l|ccc}
        \toprule
        \text{Method}                    &  \text{LPIPS} $\downarrow$ & \text{SSIM} $\uparrow$ & \text{PSNR} $\uparrow$ \\ \midrule
        \textit{Original Reconstruction} & 0.177                      &  0.896                 & 28.94           \\ \midrule
        \text{Naive Insertion}           & 0.211                      & \sbf{0.866}            & 24.48           \\
        \text{Naive Insertion + R3D2}    & \sbf{0.142}                & 0.862                  & \sbf{25.29}  \\
        \text{Naive Insertion + Difix3D} & 0.254                      & 0.806                  & 23.58   \\
        \bottomrule
    \end{tabular}
\end{table}

\subsection{Shortcomings of {\modelname}}
As discussed in \cref{sec:conclusion}, currently {\modelname} does not address the issue of temporal consistency. Because {\modelname} processes each frame independently, applying it to sequential frames in a video or across views from a multi-camera rig can result in subtle flickering or visual inconsistencies. While we believe that extending our approach to video models will resolve this, current video-based architectures are computationally intensive.

\section{Neural reconstruction details}
\subsection{Reconstruction sequence selection}
Current neural reconstruction methods struggle in extreme weather conditions, such as heavy rain/fog/snow or when the cameras experience harsh sun glares. We base our selection on filtering such scenes as well as removing scenes without any dynamic object presents. 

\subsection{Deviations from original reconstruction method}
SplatAD~\cite{hess2024splatad} employs \textit{feature splatting}, meaning that instead of rendering an image with RGB-color ($H\times W \times 3$) as vanilla gaussian splatting does, they render a feature-map  $H\times W \times f$, where $f \in \mathbf{R}^d$.
They then use a CNN-based decoder to obtain the final image $\mathcal{D} : \mathbf{R}^d \rightarrow \mathbf{R}^3$ and show that this yields better reconstruction results~\cite{hess2024splatad}.
However, as the decoder is also scene-specific, it hinders seamless transfer of objects across scenes.
We therefore disable the feature splatting and use the vanilla approach of directly rendering the RGB color of the scene.

\subsection{Partial reconstruction of 3D assets in traditional neural rendering frameworks}
\begin{figure*}[ht]
     \centering
     \begin{subfigure}{\textwidth}
         \centering
         \includegraphics[trim={150px 150px 150px 150px},clip,width=\linewidth]{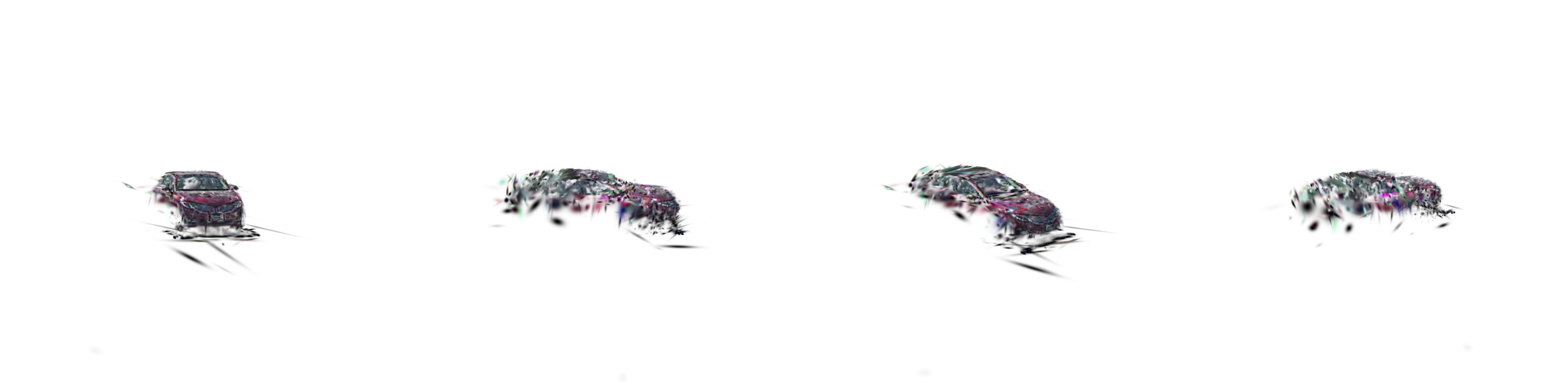}
     \end{subfigure} \\
     \begin{subfigure}{\textwidth}
         \centering
         \includegraphics[trim={150px 150px 150px 150px},clip,width=\linewidth]{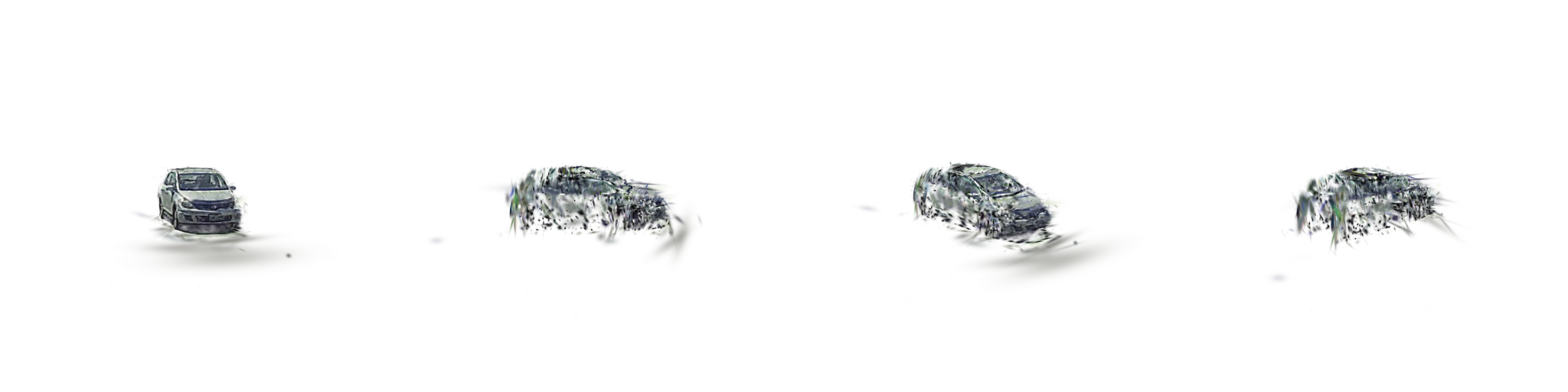}
     \end{subfigure} \\
      \begin{subfigure}{\textwidth}
         \centering
         \includegraphics[trim={150px 150px 150px 150px},clip,width=\linewidth]{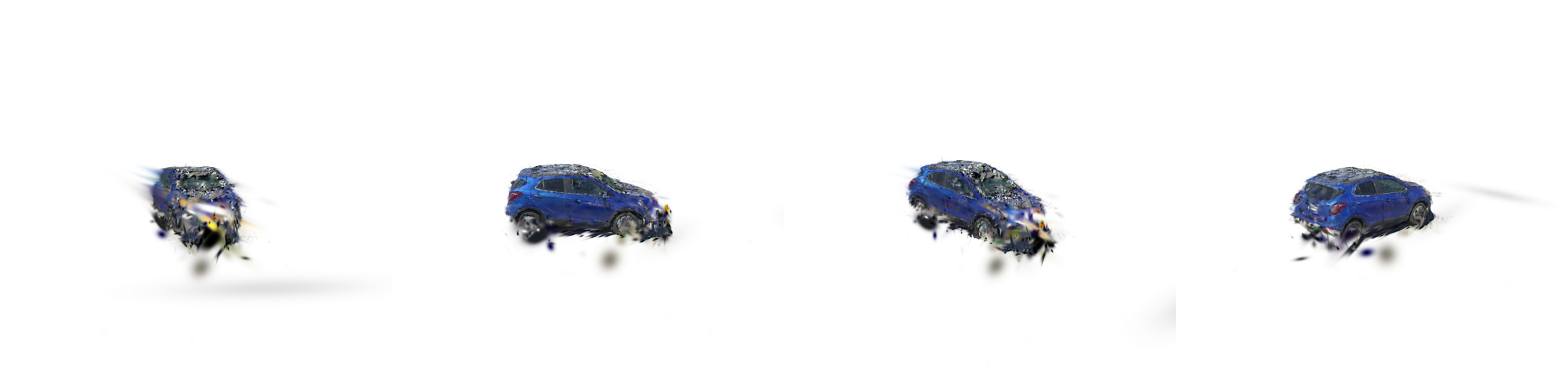}
     \end{subfigure} \\
      \begin{subfigure}{\textwidth}
         \centering
         \includegraphics[trim={150px 150px 150px 150px},clip,width=\linewidth]{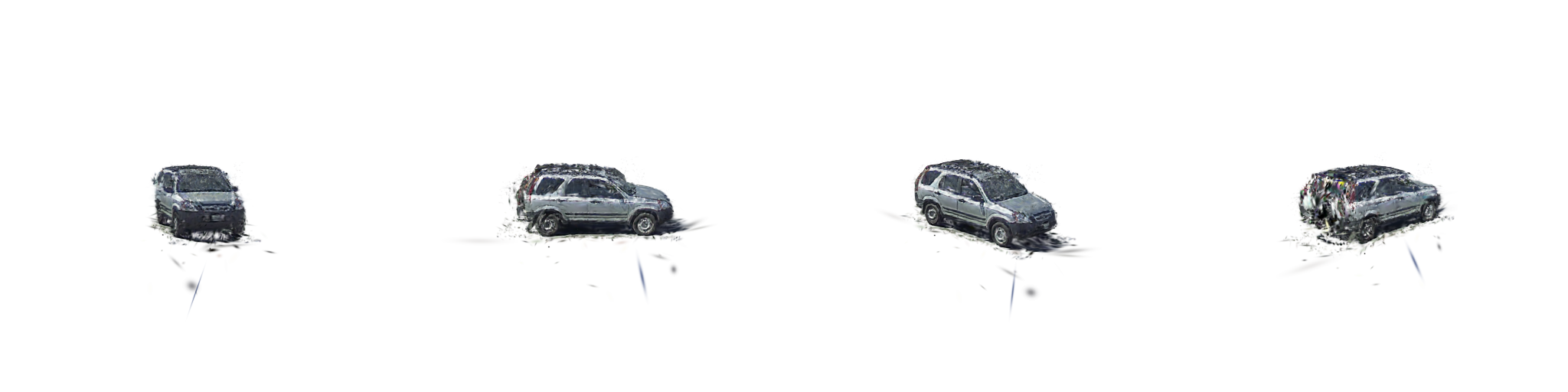}
     \end{subfigure} \\
       \begin{subfigure}{\textwidth}
         \centering
         \includegraphics[trim={150px 150px 150px 150px},clip,width=\linewidth]{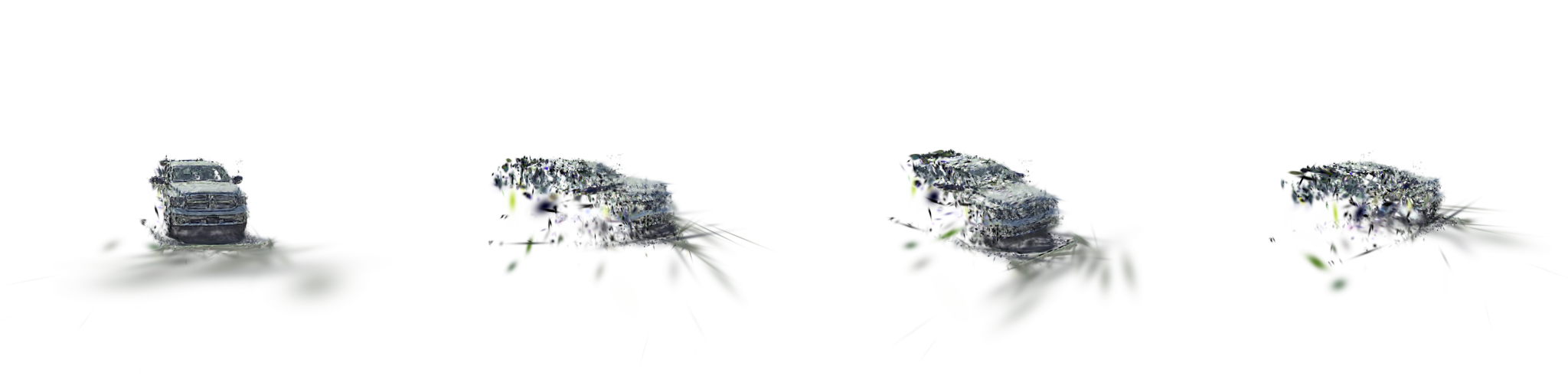}
     \end{subfigure} \\
     \caption{Multi-view images of partially reconstructed dynamic objects.}
     \label{fig:supp:partial_reconstruction}
\end{figure*}

As discussed in the main manuscript, although state-of-the-art autonomous driving (AD) scene reconstruction methods achieve impressive results, they often produce partially reconstructed objects. This limitation arises because objects in typical AD data are frequently observed from limited, often similar, viewpoints. To illustrate this issue, we present multi-view images of partially reconstructed objects in \cref{fig:supp:partial_reconstruction}. As shown, objects are generally well reconstructed from certain viewpoints but lack detail and completeness when viewed from novel directions.

\section{Dataset details}
\subsection{Asset generation with Amodal3R}
TRELLIS~\cite{xiang2024structured} enables text- and/or image-conditioned 3D asset generation. A key limitation is its assumption that conditioning images are object-centric and non-occluded. This poses a problem for in-the-wild vehicle data, where objects are often partially occluded (e.g., by other cars or barriers). To overcome this, we employ Amodal3R~\cite{wu2025amodal3r}, which facilitates image conditioning with occluded objects. Amodal3R requires three inputs: 1) the object-centric (potentially occluded) original image crop, 2) a segmentation mask of the object of interest, and 3) a foreground occlusion mask highlighting the object's occluded regions.

While the original image crop and the object's segmentation mask are readily available from WOD~\cite{waymo} and its object-assets extension (WOD-OA)~\cite{shen2023gina3d}, the foreground occlusion mask requires further processing. We generate this mask using pix2gestalt~\cite{ozguroglu2024pix2gestalt}, a method that recovers an object's complete shape from a partially occluded view. Specifically, pix2gestalt produces an amodal (whole object) mask. Because we observed that pix2gestalt tends to slightly overestimate the object's size, we first erode its generated amodal mask using a kernel of size $k=10$. The foreground occlusion mask is then obtained by subtracting the original object segmentation mask (from WOD-OA) from this eroded amodal mask.

Amodal3R employs a dual-diffusion process: the first generates the object's structure, and the second, conditioned on this structure, generates its features (which are used to generate properties like color and opacity when decoding Gaussian primitives). We use $n_{\text{steps}} = 250$ for each process, keeping other parameters at their default values. To condition on multiple images of the target object, we utilize Amodal3R's \textit{stochastic multi-sampling}, which randomly samples a new conditioning image at each diffusion timestep. Further details are available in the original publication~\cite{wu2025amodal3r}.

\subsection{Text-to-3D generation}
\begin{figure*}[ht]
     \centering
     \begin{subfigure}{0.48\textwidth}
         \centering
         \includegraphics[trim={150px 150px 0px 150px},clip,width=\linewidth]{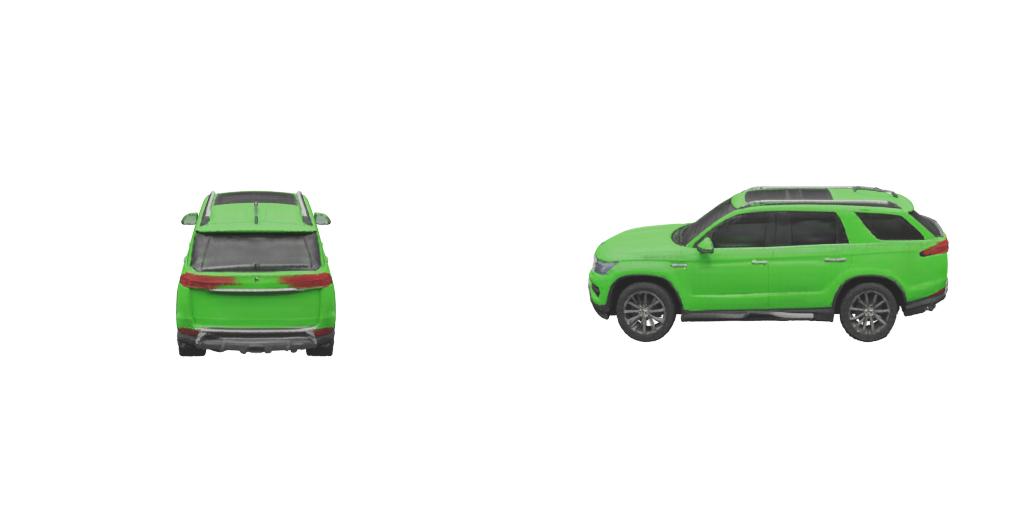}
     \end{subfigure}
     \begin{subfigure}{0.48\textwidth}
         \centering
         \includegraphics[trim={150px 150px 0px 150px},clip,width=\linewidth]{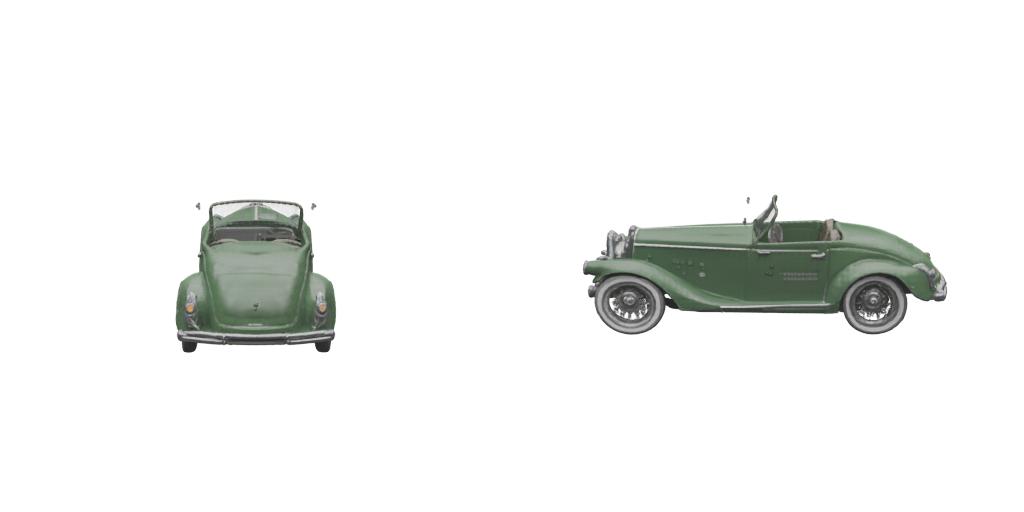}
     \end{subfigure} \\
      \begin{subfigure}{0.48\textwidth}
         \centering
         \includegraphics[trim={150px 150px 0px 150px},clip,width=\linewidth]{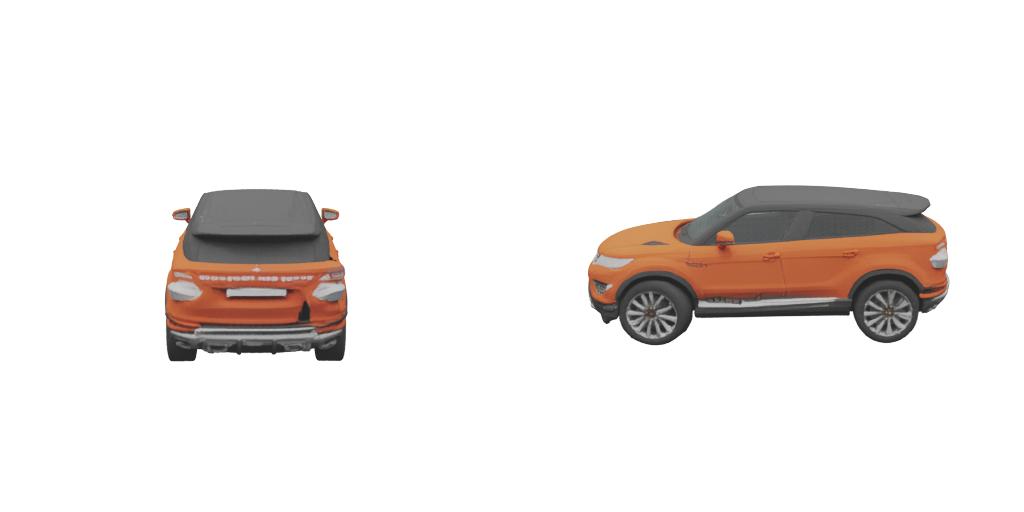}
     \end{subfigure} 
      \begin{subfigure}{0.48\textwidth}
         \centering
         \includegraphics[trim={150px 150px 0px 150px},clip,width=\linewidth]{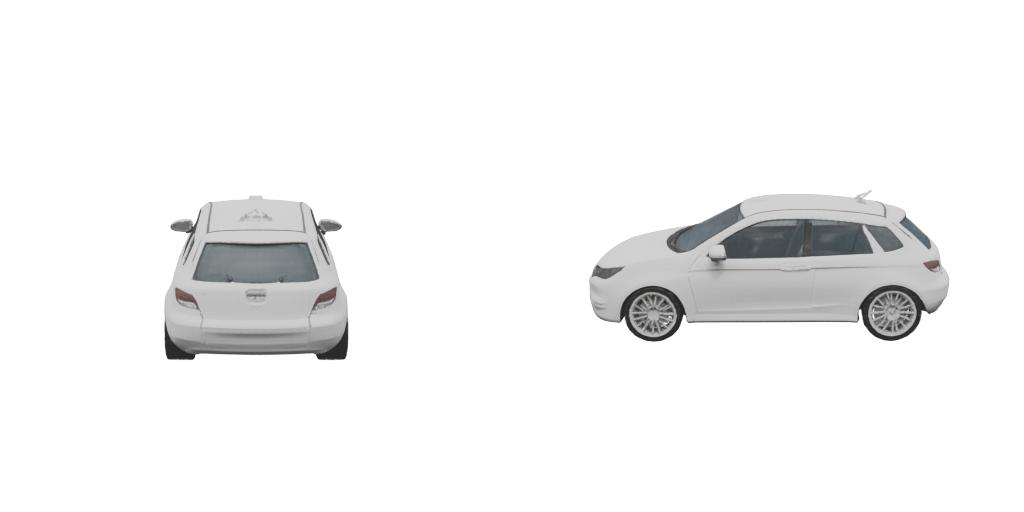}
     \end{subfigure} \\
     \begin{subfigure}{0.48\textwidth}
         \centering
         \includegraphics[trim={150px 150px 0px 150px},clip,width=\linewidth]{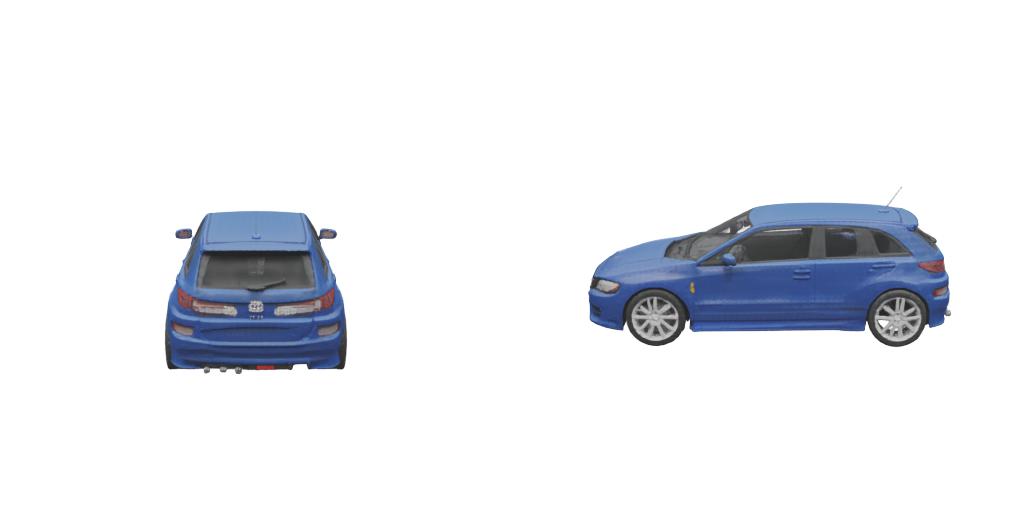}
     \end{subfigure} 
      \begin{subfigure}{0.48\textwidth}
         \centering
         \includegraphics[trim={150px 150px 0px 150px},clip,width=\linewidth]{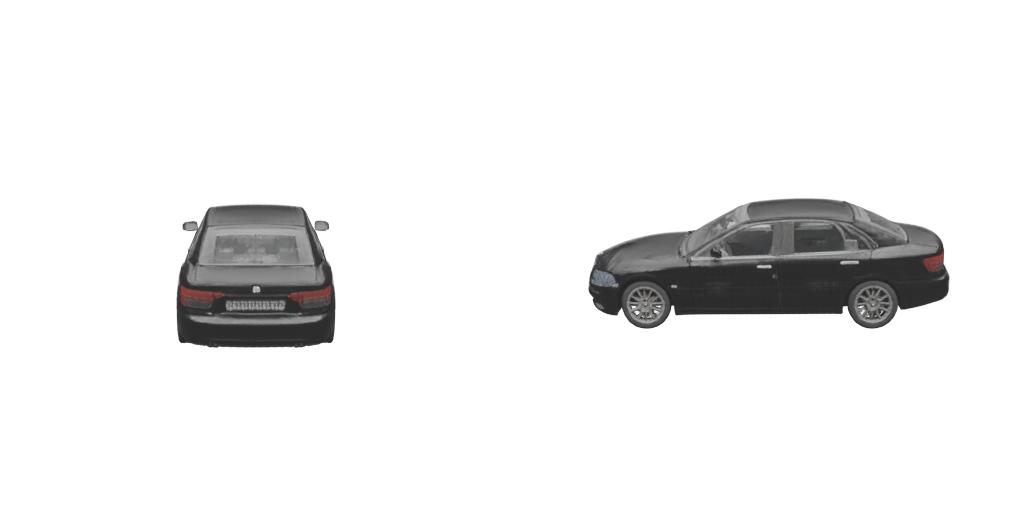}
     \end{subfigure} \\
     \caption{Example of text-to-3D assets generated with TRELLIS~\cite{xiang2024structured}.}
     \label{fig:supp:text-to-3d}
\end{figure*}

To generate our text-to-3D assets we use TRELLIS~\cite{xiang2024structured}. As text-conditioning we use a combination of randomly sampled types, styles, and colors according to the sets of prompts below. Example assets are shown in \cref{fig:supp:text-to-3d}.

For vehicle types we use:
\texttt{car}, \texttt{truck}, \texttt{van}, \texttt{SUV}, \texttt{sedan}, \texttt{coupe}, \texttt{hatchback}, \texttt{convertible}.

For styles we use:
\texttt{new}, \texttt{old}, \texttt{rusty}, \texttt{shiny}, \texttt{dusty}, \texttt{muddy}, \texttt{sporty}, \texttt{modern},  \texttt{luxury},  \texttt{vintage}.

For colors we use:
\texttt{red},  \texttt{blue},  \texttt{green},  \texttt{yellow},  \texttt{black},  \texttt{white},  \texttt{silver},  \texttt{gray}, \texttt{brown},  \texttt{orange}.  

\section{Additional qualitative examples}
\begin{figure*}[ht]
    \footnotesize
    \centering
    \begin{tabular}{@{}c@{\,}@{}c@{\,}c@{}}
    \textbf{Original} (for reference) & \textbf{Naive insertion} & \textbf{+\modelname} \\
    \includegraphics[width=0.33\textwidth]{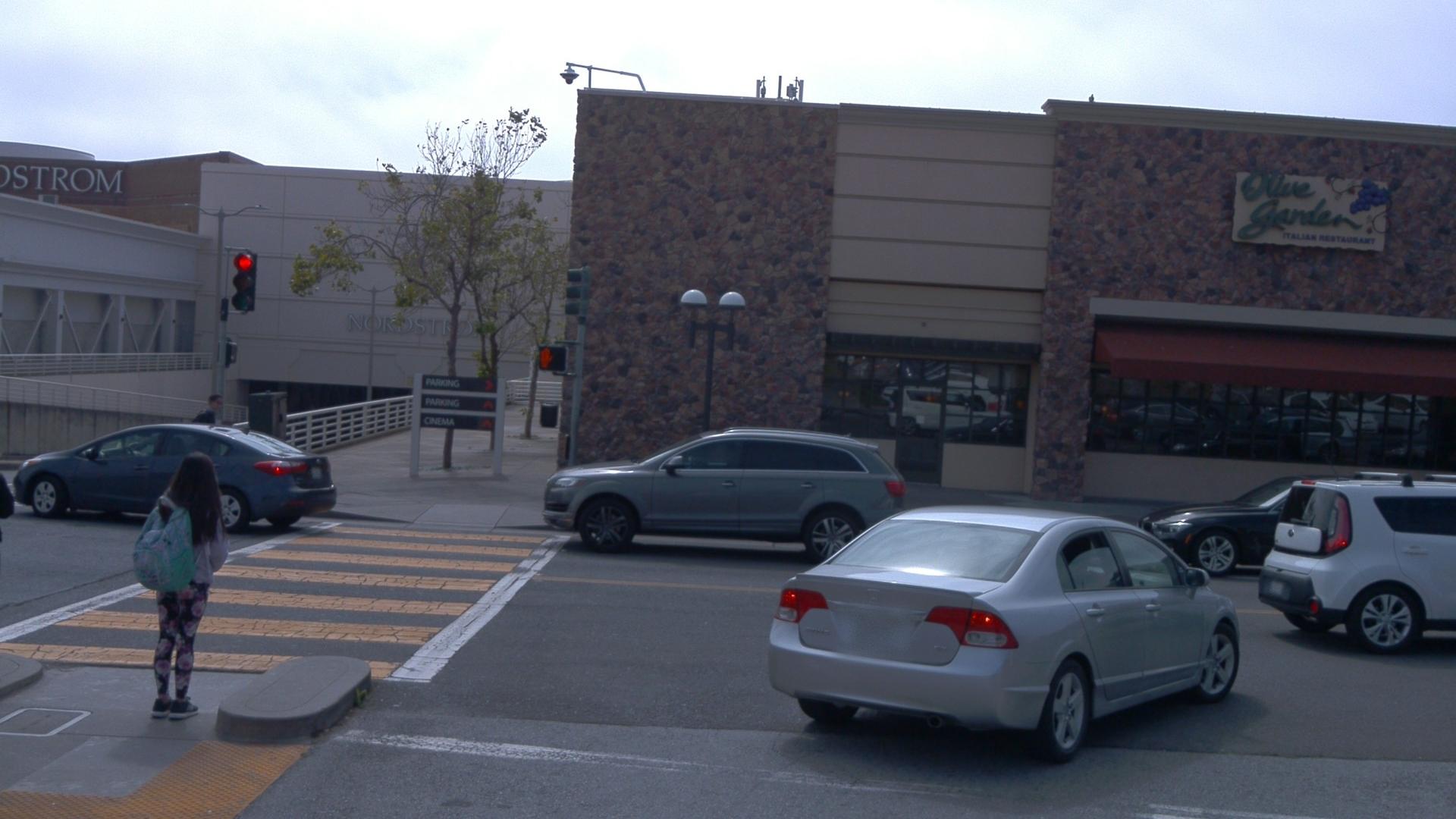} & 
    \includegraphics[width=0.33\textwidth]{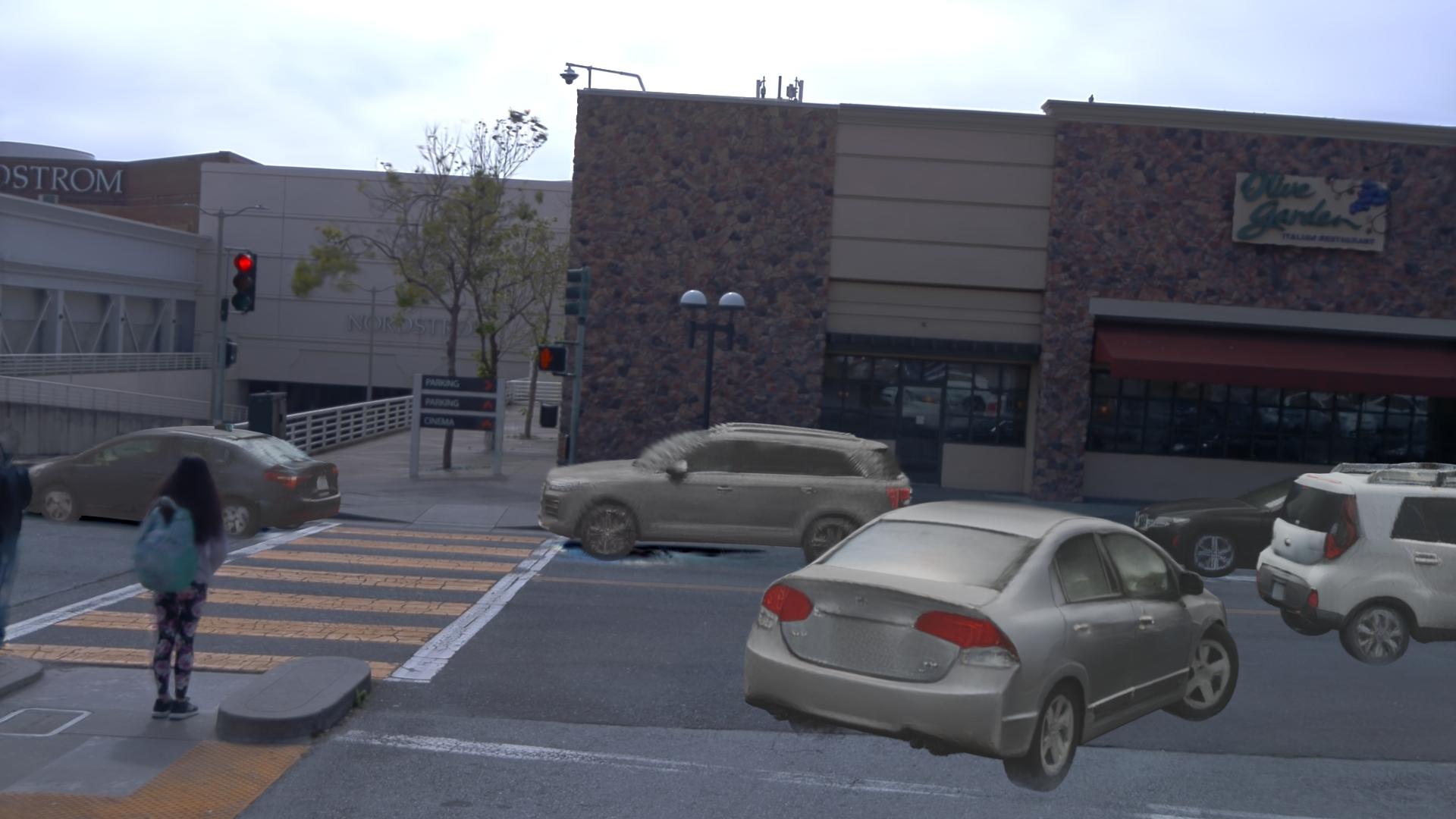} & 
    \includegraphics[width=0.33\textwidth]{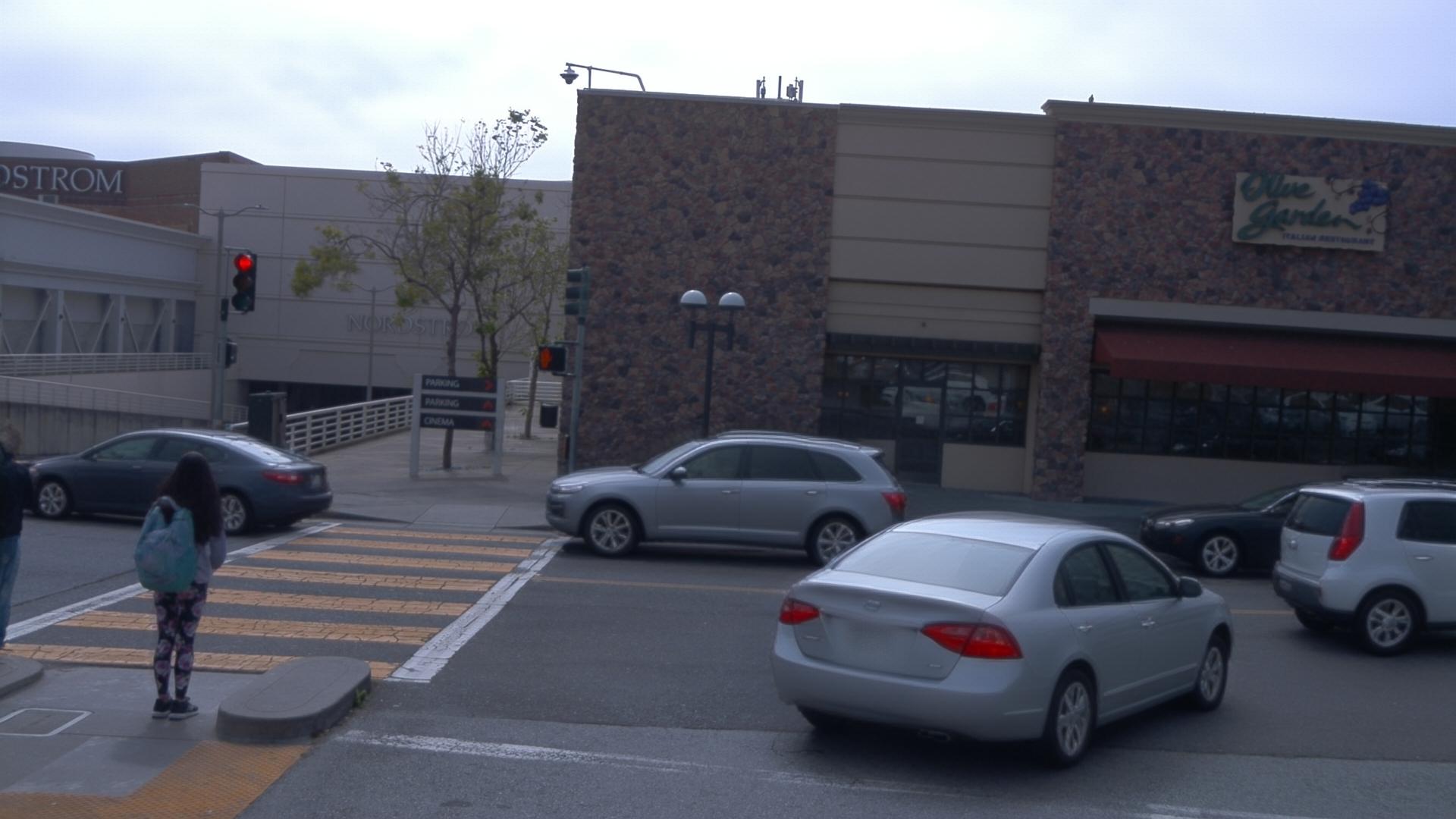} \\
    \vspace{-1.2em} \\
    \includegraphics[width=0.33\textwidth]{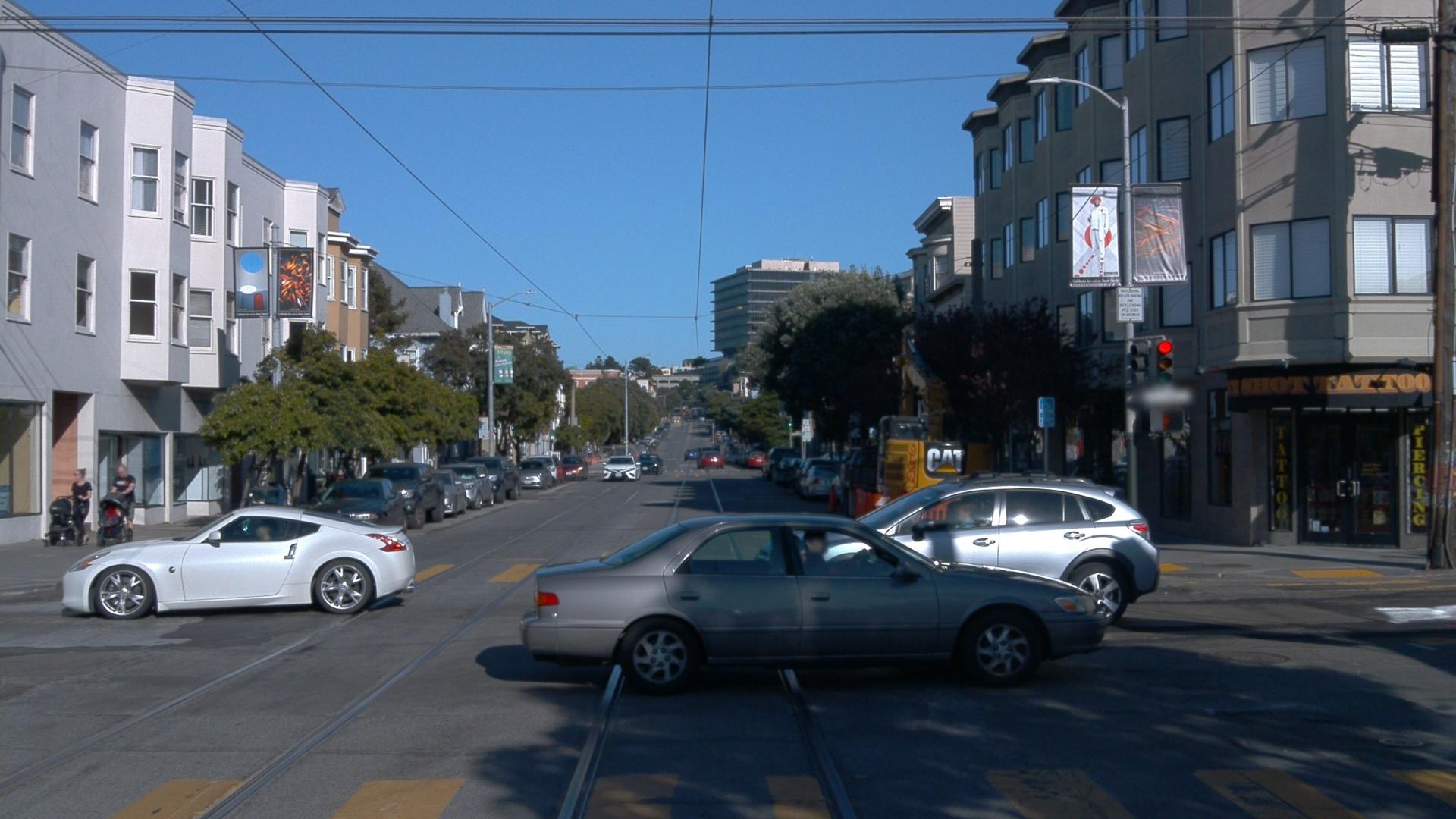} & 
    \includegraphics[width=0.33\textwidth]{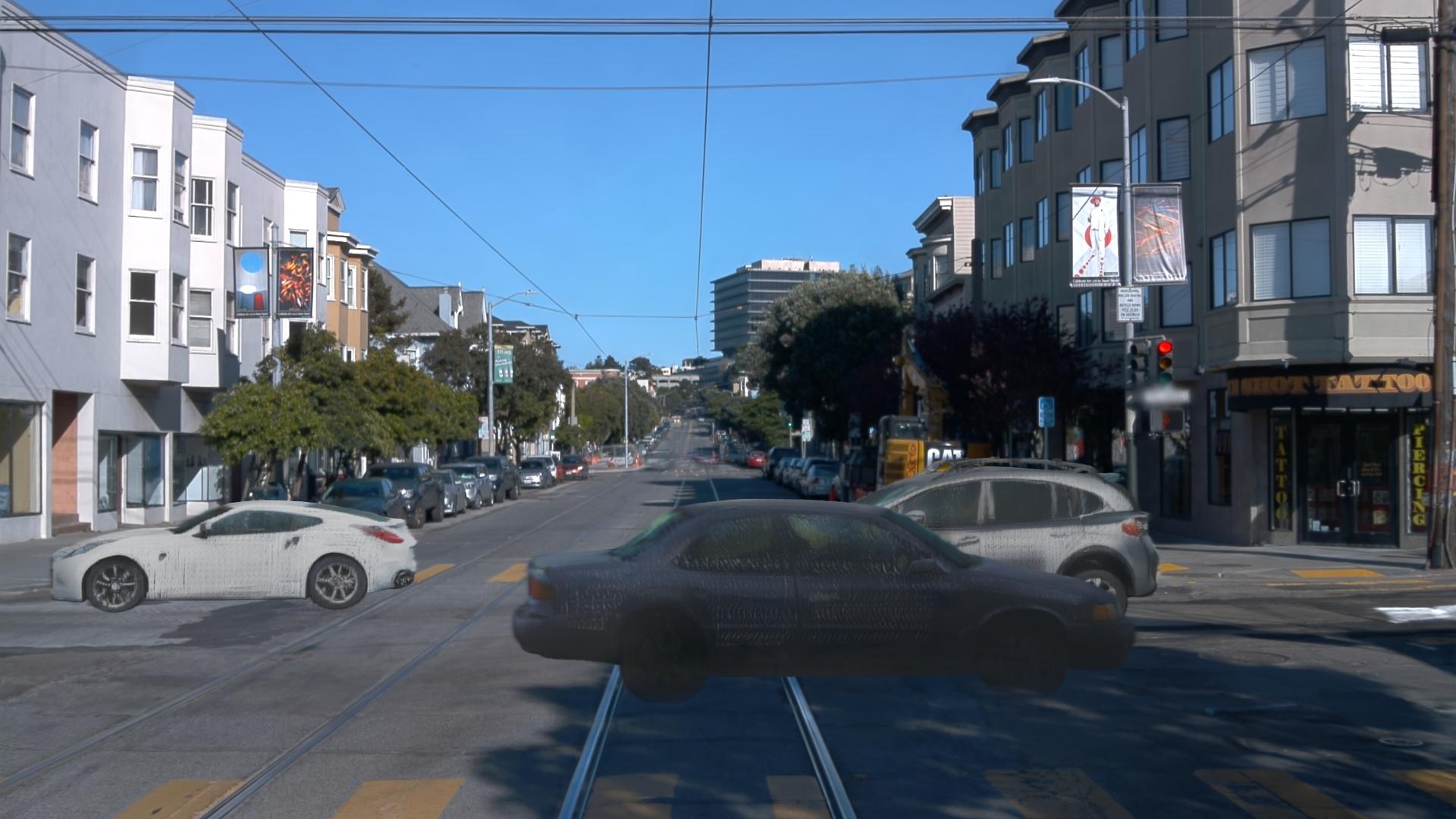} & 
    \includegraphics[width=0.33\textwidth]{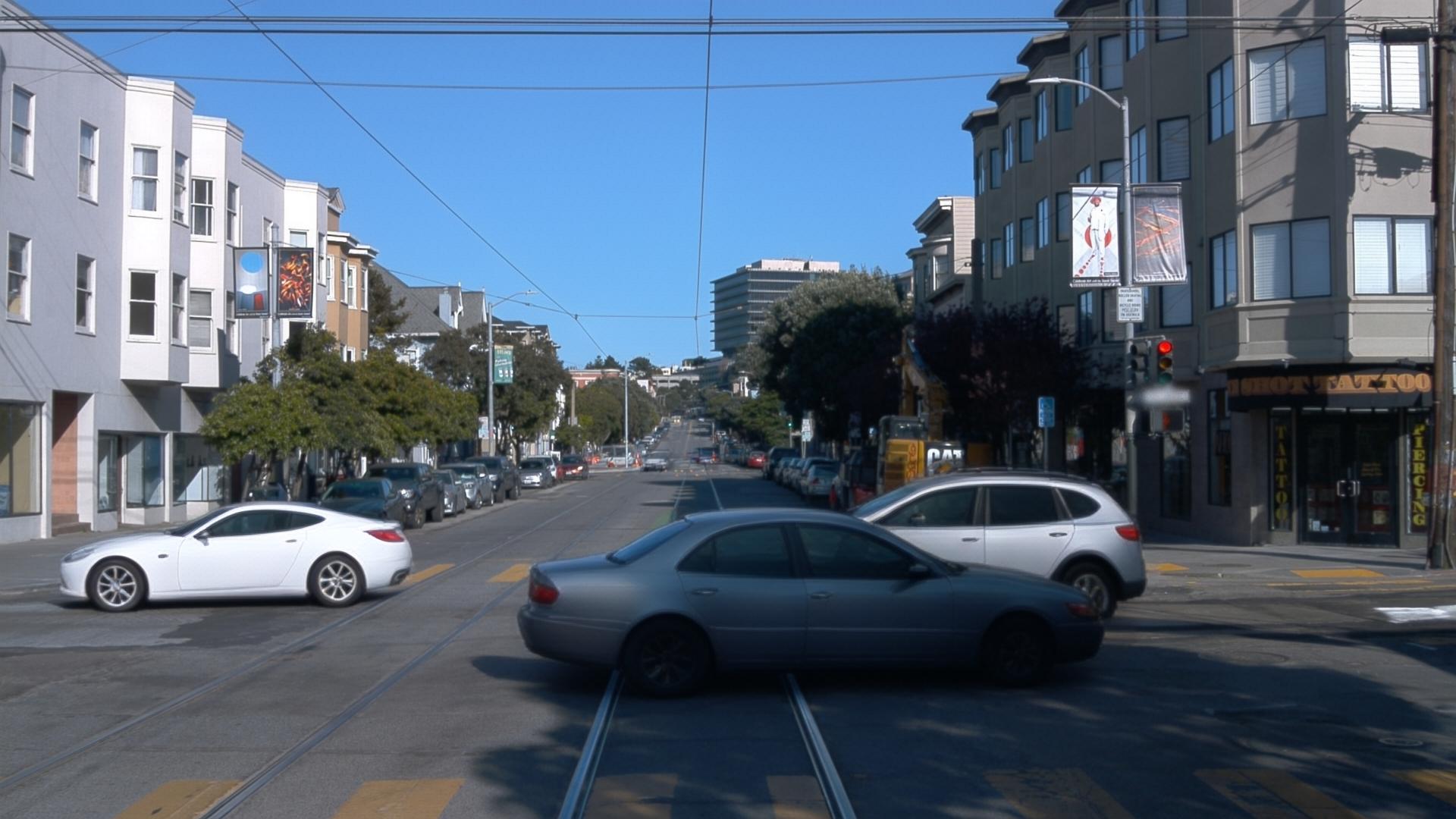} \\
    \vspace{-1.2em} \\
    \includegraphics[width=0.33\textwidth]{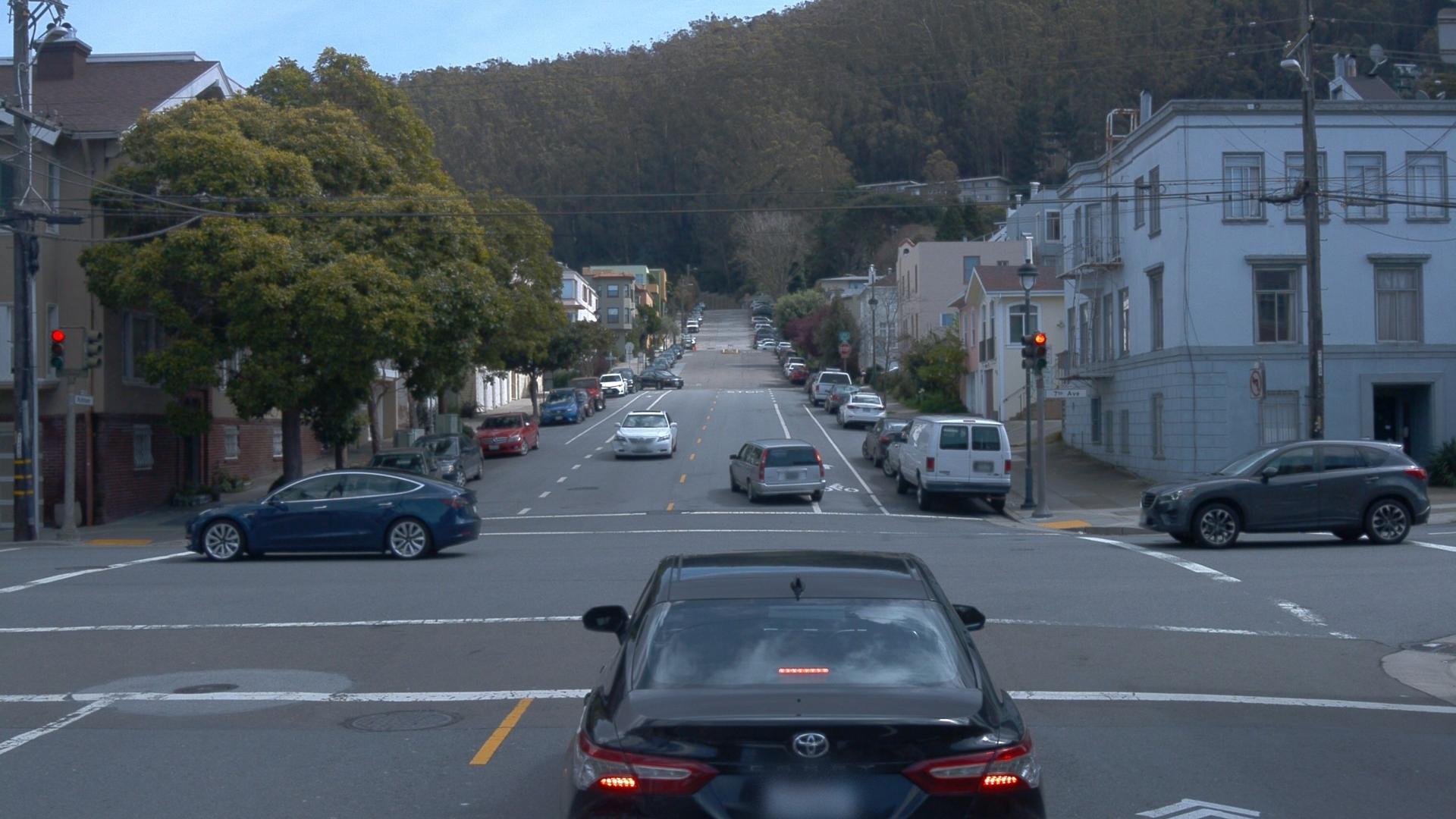} & 
    \includegraphics[width=0.33\textwidth]{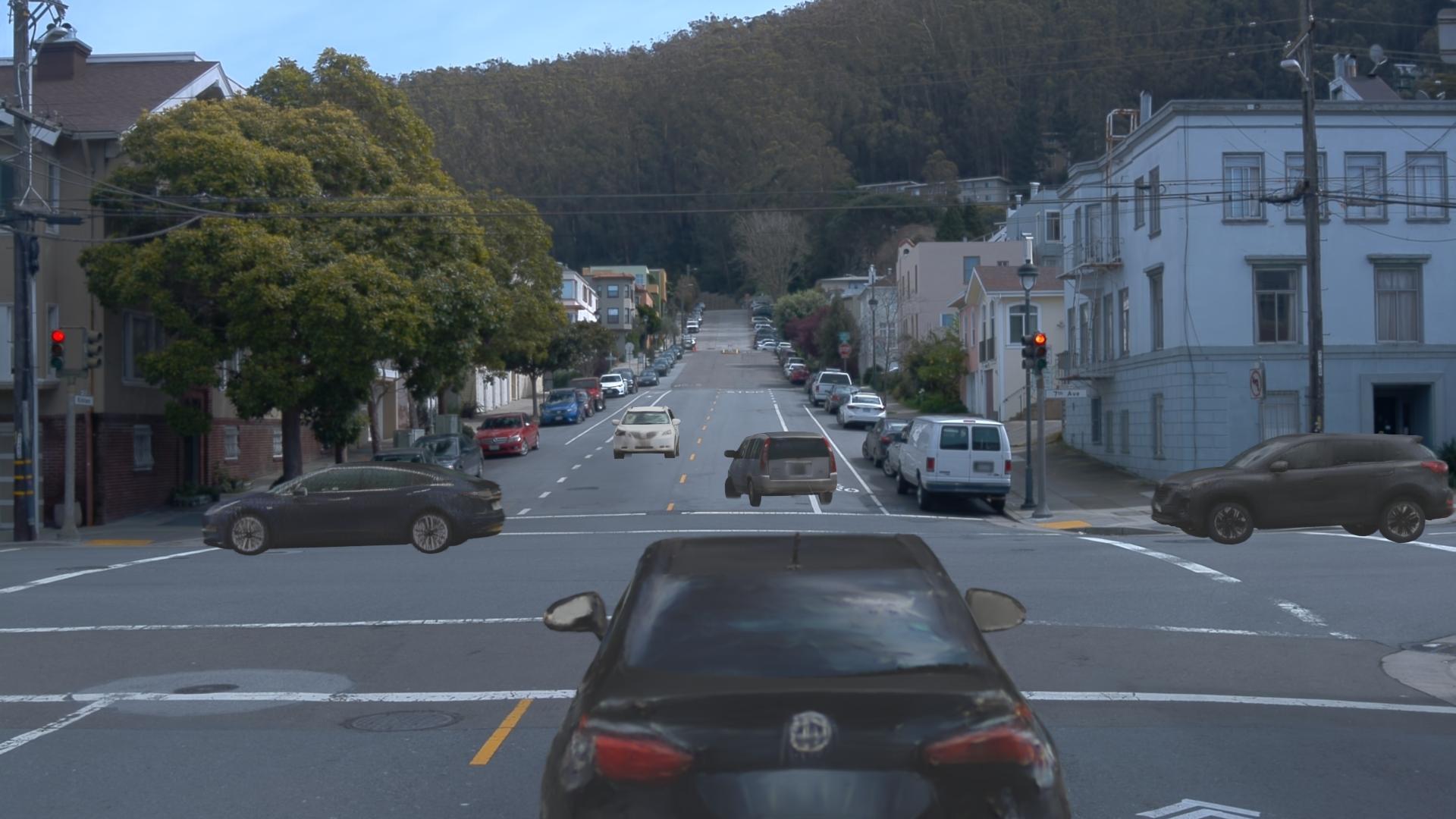} & 
    \includegraphics[width=0.33\textwidth]{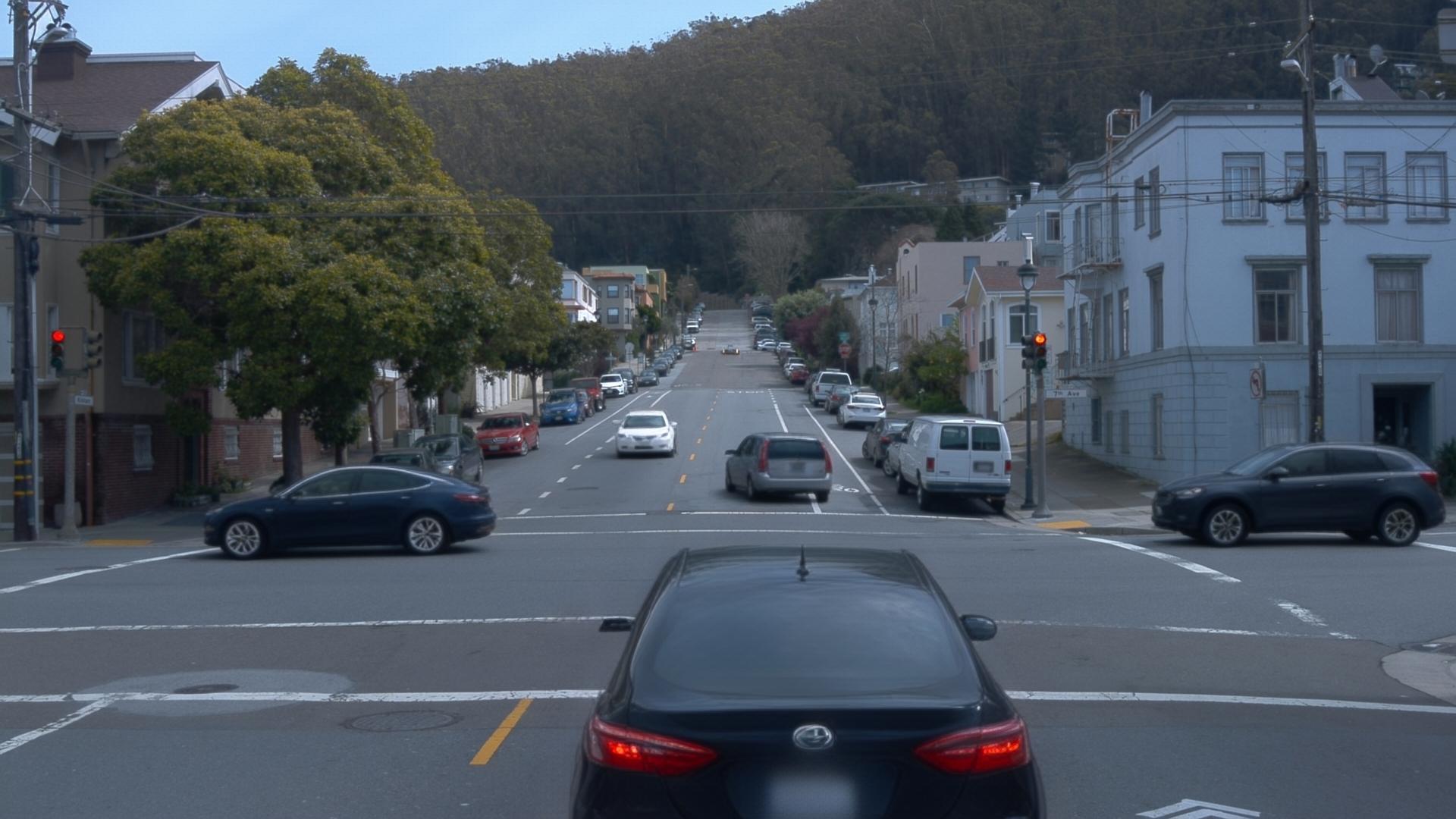} \\
    \vspace{-1.2em} \\
    \includegraphics[width=0.33\textwidth]{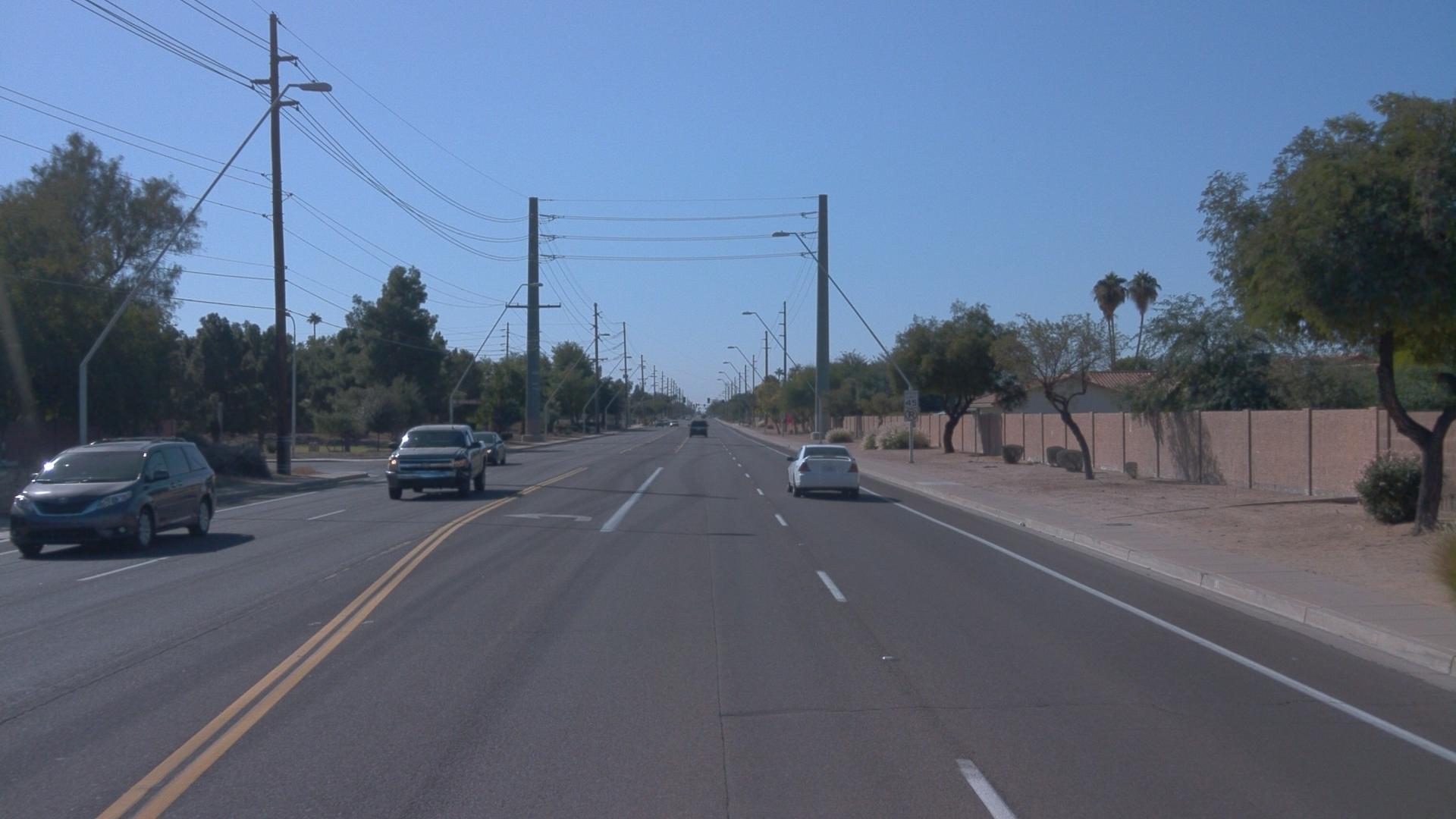} & 
    \includegraphics[width=0.33\textwidth]{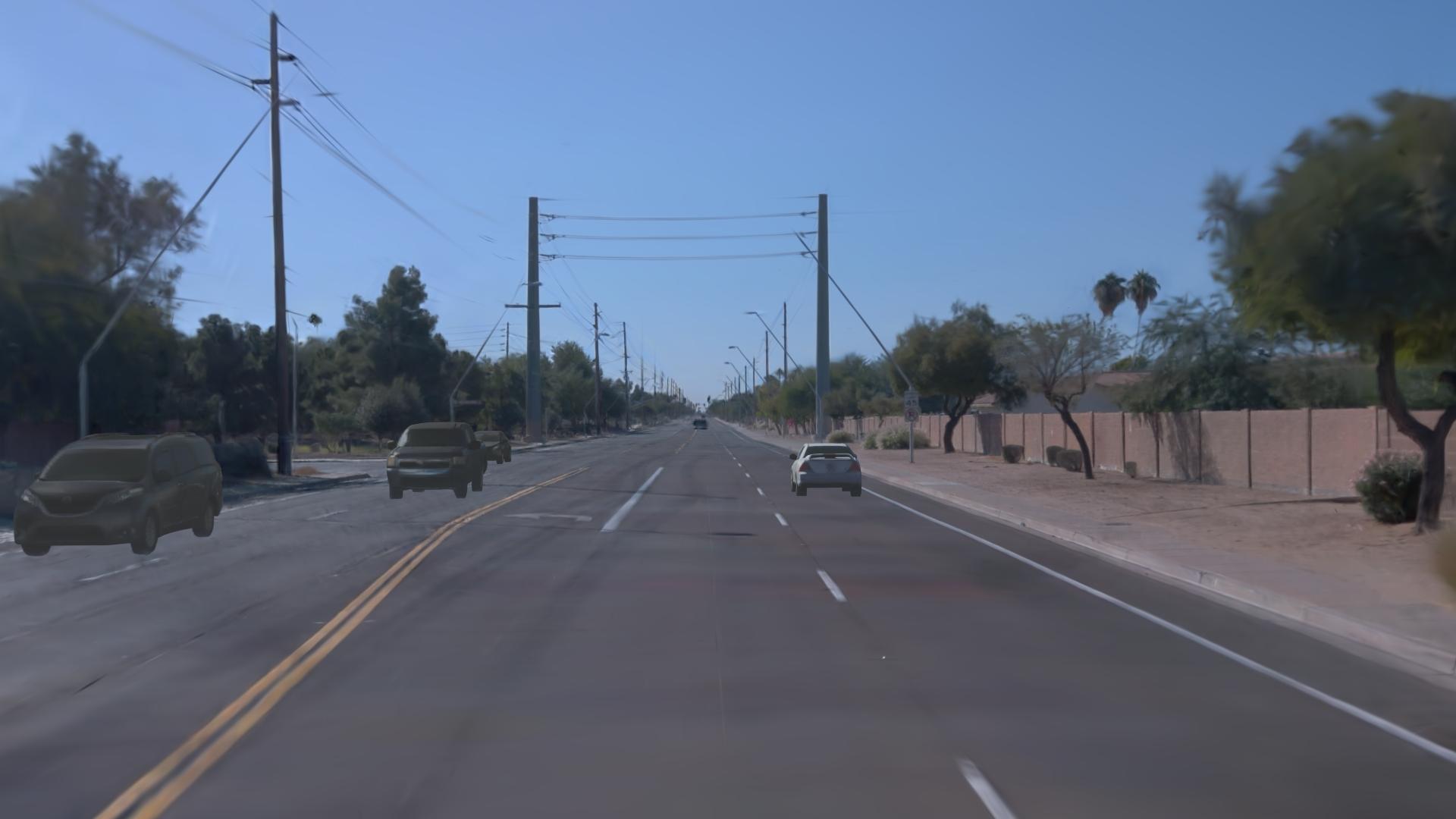} & 
    \includegraphics[width=0.33\textwidth]{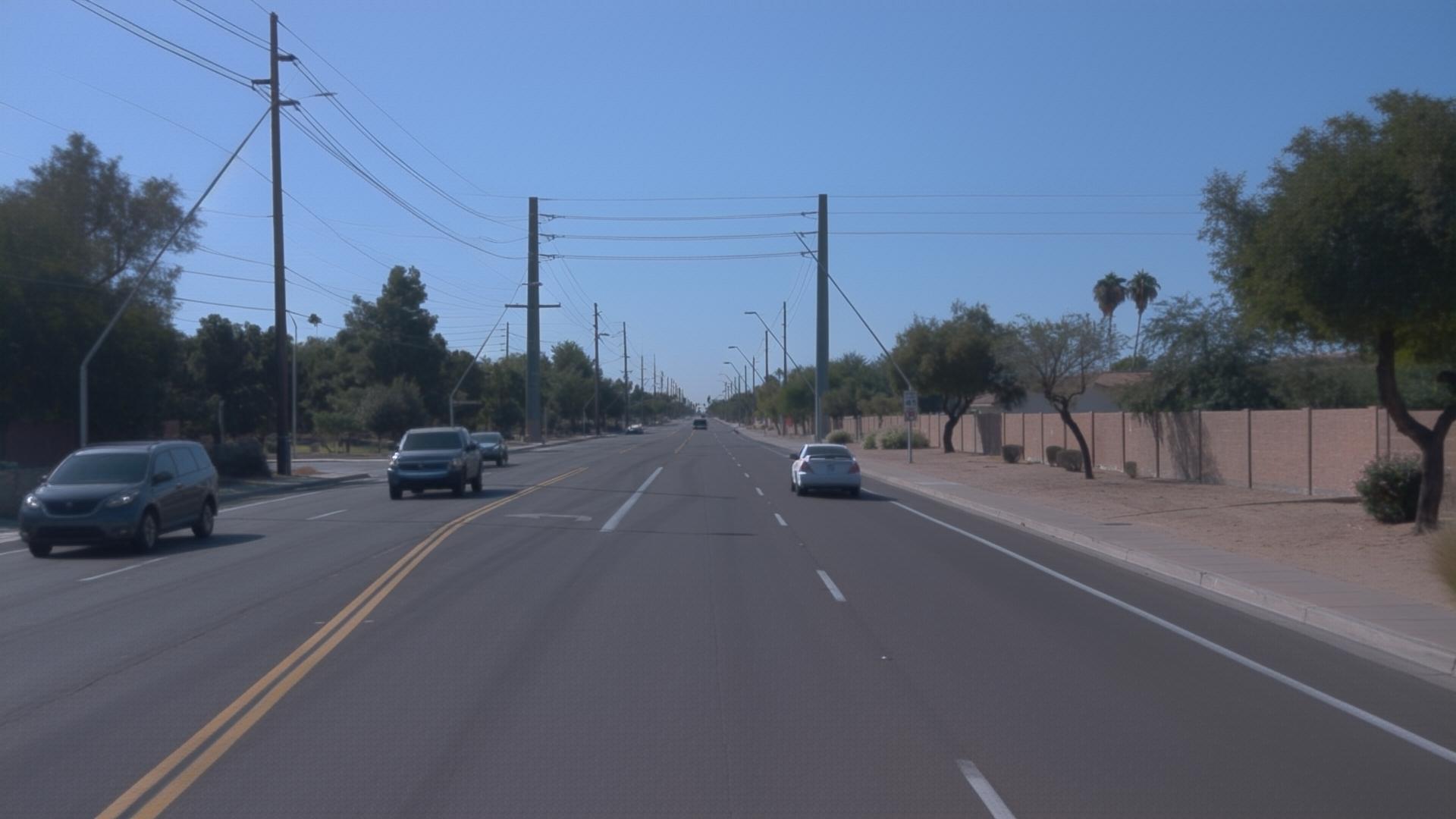} \\
    \vspace{-1.2em} \\
    \includegraphics[width=0.33\textwidth]{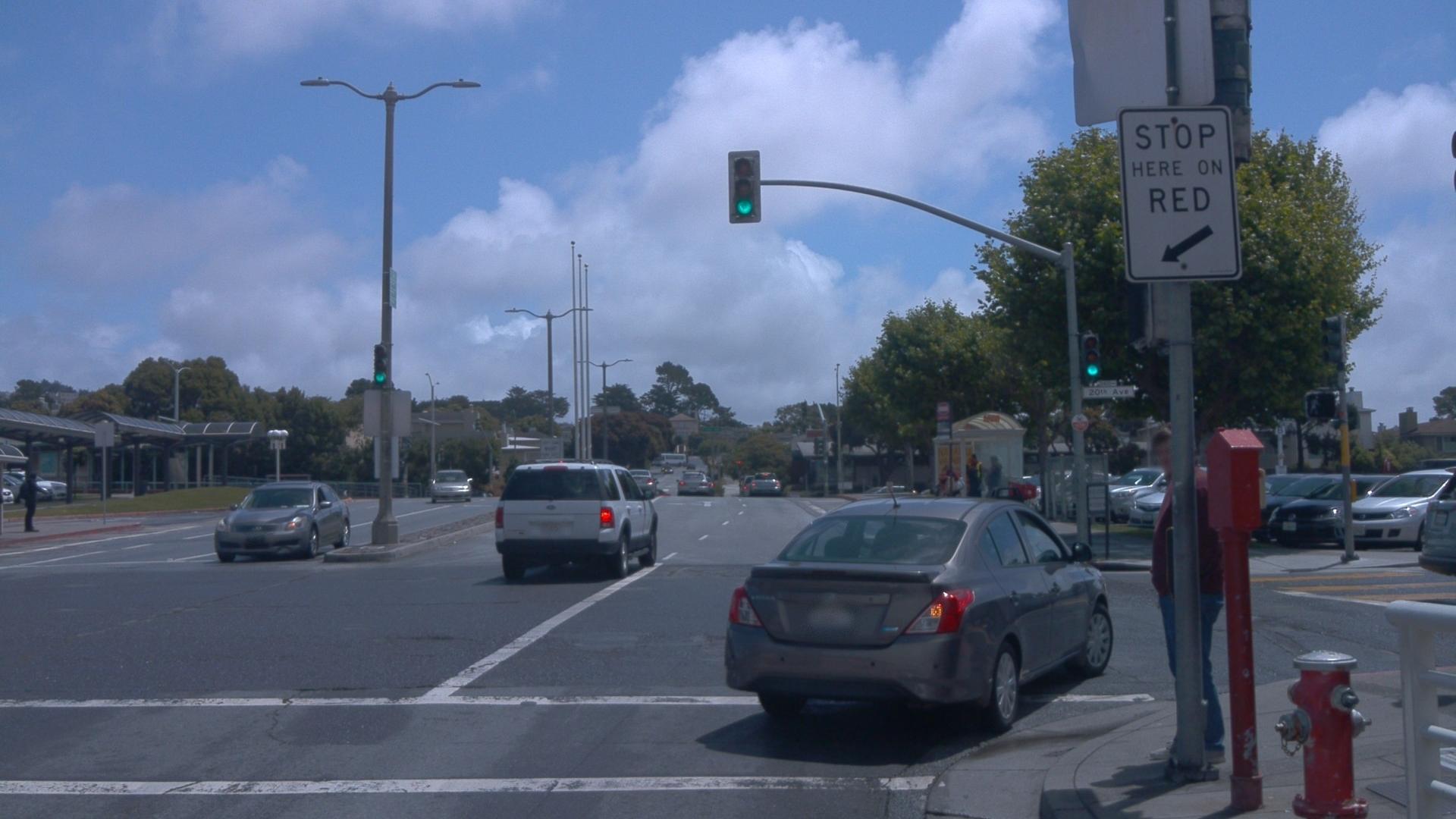} & 
    \includegraphics[width=0.33\textwidth]{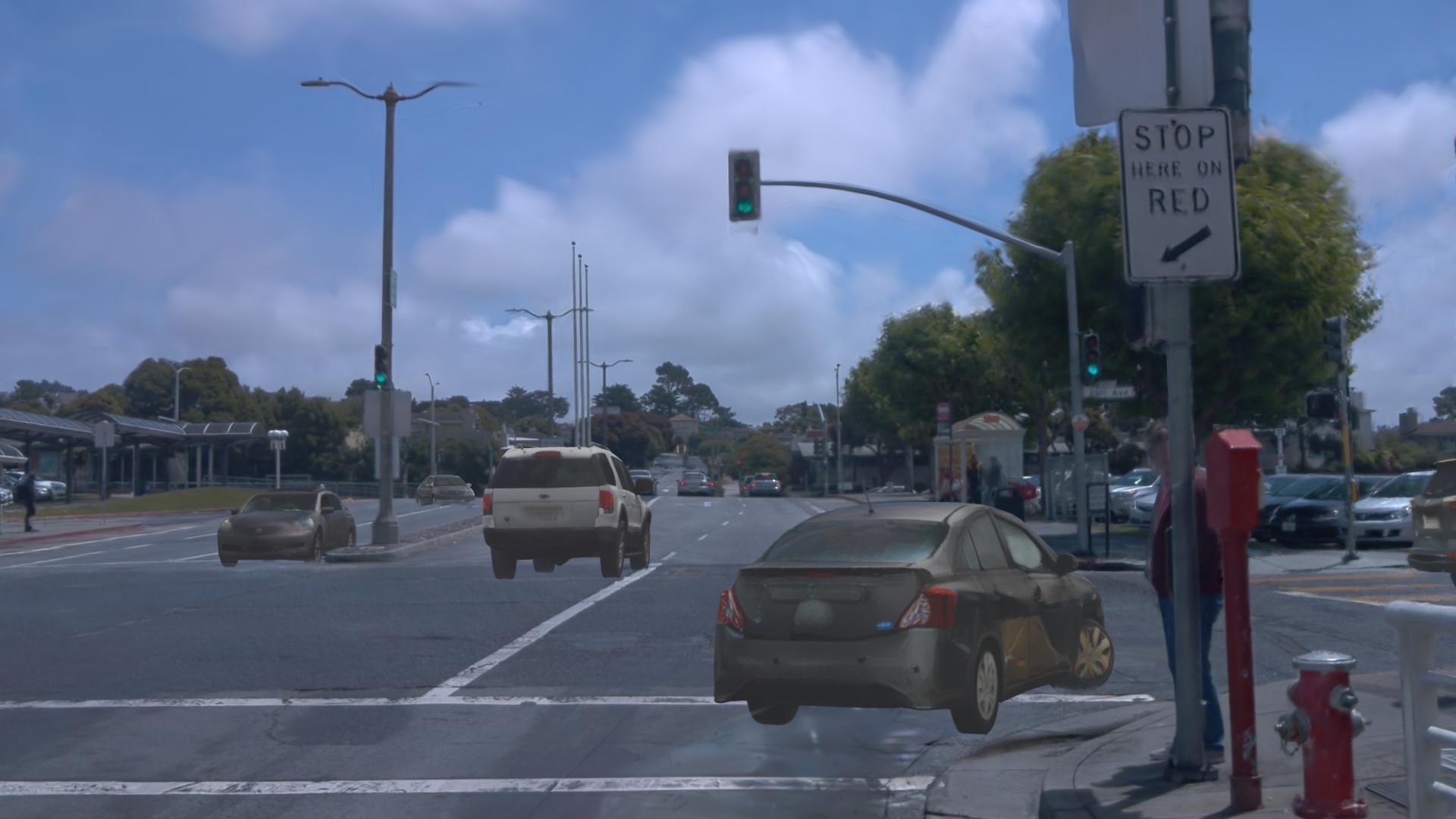} & 
    \includegraphics[width=0.33\textwidth]{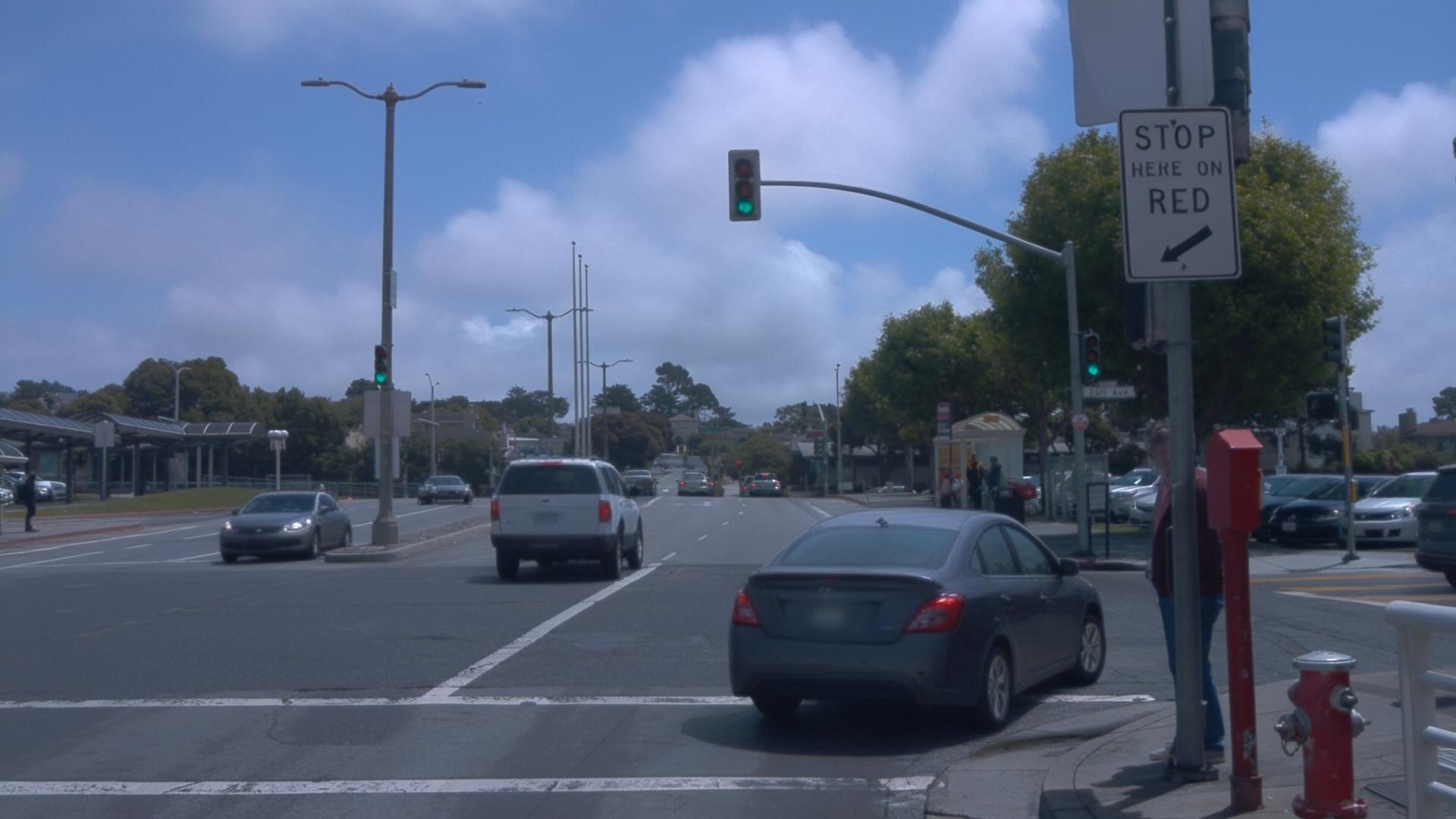} \\
    
    \end{tabular}

    \vspace{1em}
    \includegraphics[width=1.0\textwidth]{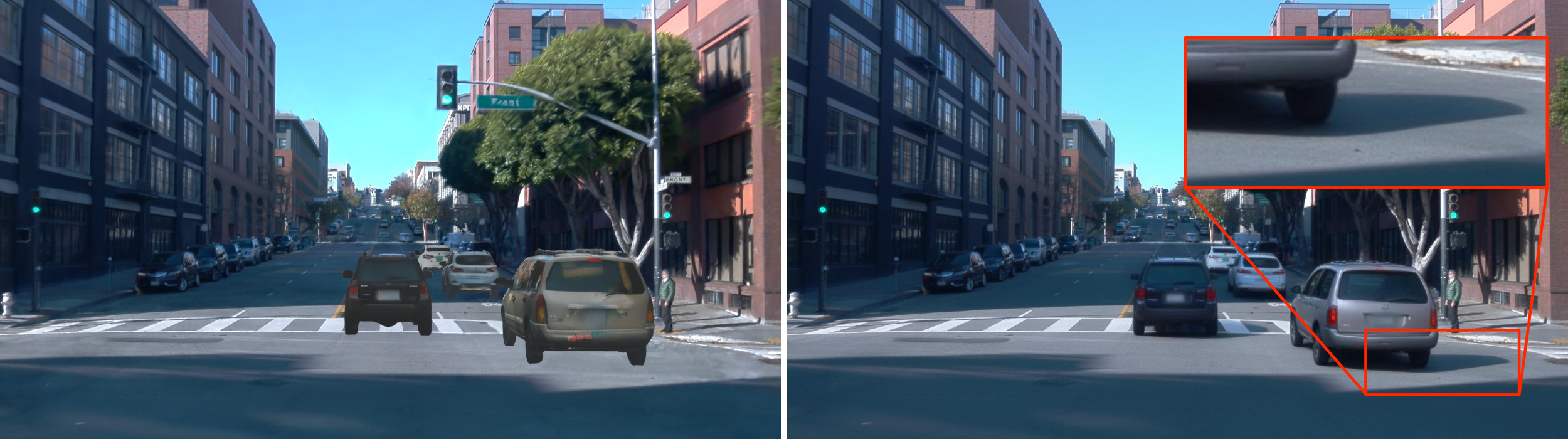} \\

    \caption{Samples of {\modelname} applied on reinserted actors.}
    \label{fig:supp:qualitative}
\end{figure*}

We show additional qualitative examples in \cref{fig:supp:qualitative} highlighting the capabilities of {\modelname}.

\end{document}